\pdfoutput=1

\documentclass[11pt]{article}

\usepackage[preprint]{acl}

\usepackage{amsmath,amsfonts,bm}

\def\eqref#1{equation~\ref{#1}}

\def\1{\bm{1}}

\DeclareMathAlphabet{\mathsfit}{\encodingdefault}{\sfdefault}{m}{sl}
\SetMathAlphabet{\mathsfit}{bold}{\encodingdefault}{\sfdefault}{bx}{n}

\usepackage{hyperref}
\usepackage{url}
\usepackage[utf8]{inputenc} 
\usepackage[T1]{fontenc}    
\usepackage{hyperref}       
\usepackage{url}            
\usepackage{booktabs}       
\usepackage{amsfonts}       
\usepackage{nicefrac}       
\usepackage{microtype}      
\usepackage{multirow}

\usepackage{graphicx}
\usepackage{booktabs}
\usepackage{threeparttable}
\usepackage{wrapfig}
\usepackage{enumitem}
\usepackage{colortbl}
\usepackage[accsupp]{axessibility}  
\usepackage{CJKutf8}

\usepackage{color}

\usepackage{stfloats}
\usepackage{float}
\usepackage{algorithm}
\usepackage{algpseudocode}
\usepackage{times}

\title{SR-LLM: Rethinking the Structured Representation in Large Language Model}

\author{Jiahuan Zhang\textsuperscript{1,2,3}\thanks{~~Visiting at Westlake University}, 
Tianheng Wang\textsuperscript{1}, 
Hanqing Wu\textsuperscript{2},  
Ziyi Huang\textsuperscript{1,4}\footnotemark[1],  
Yulong Wu\textsuperscript{5},\\
\textbf{Dongbai Chen\textsuperscript{2}},  
\textbf{Linfeng Song\textsuperscript{6}},  
\textbf{Yue Zhang\textsuperscript{1}},  
\textbf{Guozheng Rao\textsuperscript{3}\thanks{~~Corresponding author}} ,
\textbf{Kaicheng Yu\textsuperscript{1,2}\footnotemark[2]}\\
    \textsuperscript{1} Westlake University
    \textsuperscript{2} KMind Technology Co., Ltd.     \textsuperscript{3} Tianjin University\\
    \textsuperscript{4} Beijing Jiaotong University, Weihai
    \textsuperscript{5} University of Toronto
    \textsuperscript{6} Tencent AI Lab\\
    \texttt{\{zhangjiahuan78, kyu\}@westlake.edu.cn}
}

\begin{document}

\maketitle

\begin{abstract}
Structured representations, exemplified by Abstract Meaning Representation~(AMR), have long been pivotal in computational linguistics. However, their role remains ambiguous in the Large Language Models (LLMs) era. Initial attempts to integrate structured representation into LLMs via a zero-shot setting yielded inferior performance. We hypothesize that such a decline stems from the structure information being passed into LLMs in a code format unfamiliar to LLMs' training corpora. Consequently, we propose SR-LLM, an innovative framework with two settings to explore a superior way of integrating structured representation with LLMs from training-free and training-dependent perspectives. The former integrates structural information through natural language descriptions in LLM prompts, whereas its counterpart augments the model's inference capability through fine-tuning on linguistically described structured representations. Performance improvements were observed in widely downstream datasets, with particularly notable gains of 3.17\% and 12.38\% in PAWS. To the best of our knowledge, this work represents the pioneering demonstration that leveraging structural representations can substantially enhance LLMs' inference capability. We hope that our work sheds light and encourages future research to enhance the reasoning and interoperability of LLMs by structure data. 

\end{abstract}

\section{Introduction}
\begin{figure}[!ht]
\centering
\includegraphics[width=1\linewidth]{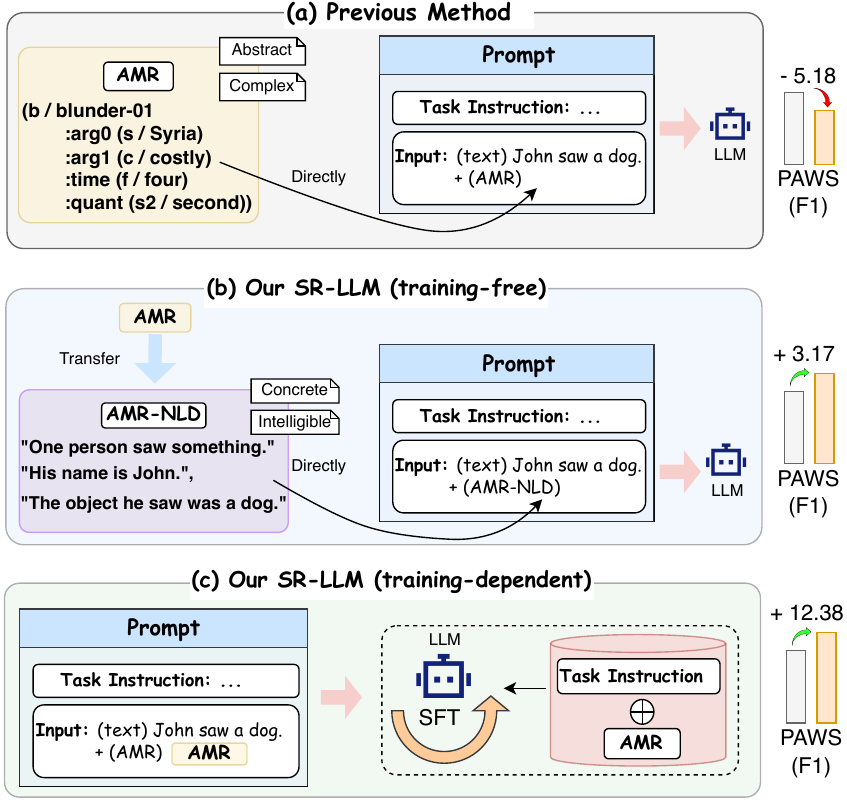}
\vspace{-0.6cm}
\caption{We propose two novel AMR integration approaches: a training-free method using natural language descriptions and a training-dependent fine-tuning paradigm. Evaluation on PAWS shows +3.17\% and +12.38\% improvements respectively, contrasting with the -5.18\% decline in conventional code-format methods.}
\vspace{-0.4cm}
\label{fig:teaser}
\end{figure}

\begin{figure}
\centering
\includegraphics[width=1\linewidth]{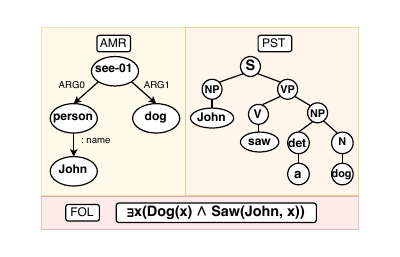}
\caption{The AMR, PST, and FOL of the sentence ``John saw a dog''.}
\label{fig:apf}
\vspace{-0.2in}
\end{figure}

Structured representations (SR), manifested in Abstract Meaning Representation (AMR)~\citep{damonte2016incremental, knight2020abstract, ramirez2024natural}, Parse Syntax Trees (PST)~\citep{sachan2020syntax}, and First-Order Logic (FOL)~\citep{barwise1977introduction}, have been fundamental to NLP~\citep{manning1999foundations, collobert2011natural}, serving as sophisticated frameworks for capturing semantic relationships and linguistic structures~\citep{banarescu2013abstract, wang2015transition}. An example of AMR, PST, and FOL is depicted in Figure~\ref{fig:apf}. 

In the era of LLMs, the paradigm for optimal SR integration remains an open research challenge. Despite LLMs' capabilities, direct integration of SR into prompts, as illustrated in Figure~\ref{fig:teaser}, has proven counterproductive~\citep{jin2024analyzing}. We posit that this performance degradation stems from LLMs' inherent limitations in processing structured representations, where direct exposure to complex linguistic structures impedes rather than enhances their reasoning process.

To address the aforementioned challenges and effectively leverage SR in LLMs, we introduce SR-LLM, a comprehensive framework with dual configurations for structural knowledge integration. The training-free approach transforms SR into natural language descriptions (SR-NLD), enhancing prompt comprehension by reformulating structured information into semantically rich, accessible formats that facilitate nuanced reasoning and reduce ambiguity. Complementarity, the training-dependent paradigm employs supervised fine-tuning on task-specific SR datasets (termed Gen-SR) to establish robust SR-task connections through iterative exposure to structured data, enabling the model to develop sophisticated internal representations and leverage deep structural knowledge during inference across diverse NLP tasks.


Our empirical evaluation encompasses a comprehensive suite of NLP benchmarks, spanning diverse linguistic phenomena from paraphrase detection~\citep{mihalcea2006corpus, dolan2005automatically} and textual entailment recognition~\citep{dagan2005pascal,bowman2015large} to machine translation ~\citep{bahdanau2014neural, johnson2017google}. This diverse benchmark selection enables rigorous evaluation of our methods across the NLP spectrum. Experimental results demonstrate the superiority of our methods over existing approaches: on PAWS, while conventional method exhibits a 5.18\% performance degradation, our training-free and training-dependent approaches achieve +3.17\% and +12.38\% improvements respectively, which validating the efficacy of our structured information integration paradigm.


Our contributions are as follows:
\begin{itemize}
    \item We introduce SR-LLM, a novel framework that facilitates SR integration with LLMs through dual paradigms: training-free adaptation and supervised fine-tuning.
    \item We provide insights into how different types of SR (AMR, PST, FOL) impact LLMs performance across various tasks.
    \item To the best of our knowledge, we are the first to show that combining such SR does in fact improve LLM performance, which opens up new avenues for enhanced LLM reasoning and interoperability.
\end{itemize}

\section{Problem Definition}

This research endeavors to investigate the potential synergies between SR and LLMs, with the ultimate goal of ascertaining how their seamless integration can augment the efficacy and proficiency of LLMs in a wide array of NLP tasks. 

Given a natural language input sequence \( X = (x_1, x_2, \ldots, x_n) \), where \( x_i \in V \) represents a token drawn from the vocabulary \( V \), we also introduce the structured representation \( Z \). \( Z \) serves as auxiliary information derived from \( X \) and can take various forms, such as AMR, PST, or FOL. These SRs capture semantic, syntactic, or logical information and provide complementary insights to natural language understanding.

The task involves generating an output sequence \( Y = (y_1, y_2, \ldots, y_m) \), where each \( y_i \) belongs to either the target vocabulary or a structured semantic output space. This transformation is performed by a model \( f \), defined as:
    \begin{equation}
        Y = f(X, Z)
    \end{equation}
Here, \( f \) specifies how \( X \) and \( Z \) are utilized to complete a specific task by integrating natural language input with its structured representation.

The primary goal of this research is to optimize the definition of \( f \) to achieve the most effective use of \( X \) and \( Z \), thereby maximizing task performance. Specifically, the objective is to identify the optimal model \( f^* \) that maximizes the evaluation metric \( P(\cdot) \), such as accuracy or F1 score:
    \begin{equation}
f^* = \underset{f}{\text{arg max}} \, P(f(X, Z))
    \end{equation}
\begin{figure*}[t!]
\centering
\vspace{-0.2in}
\includegraphics[width=1\linewidth]{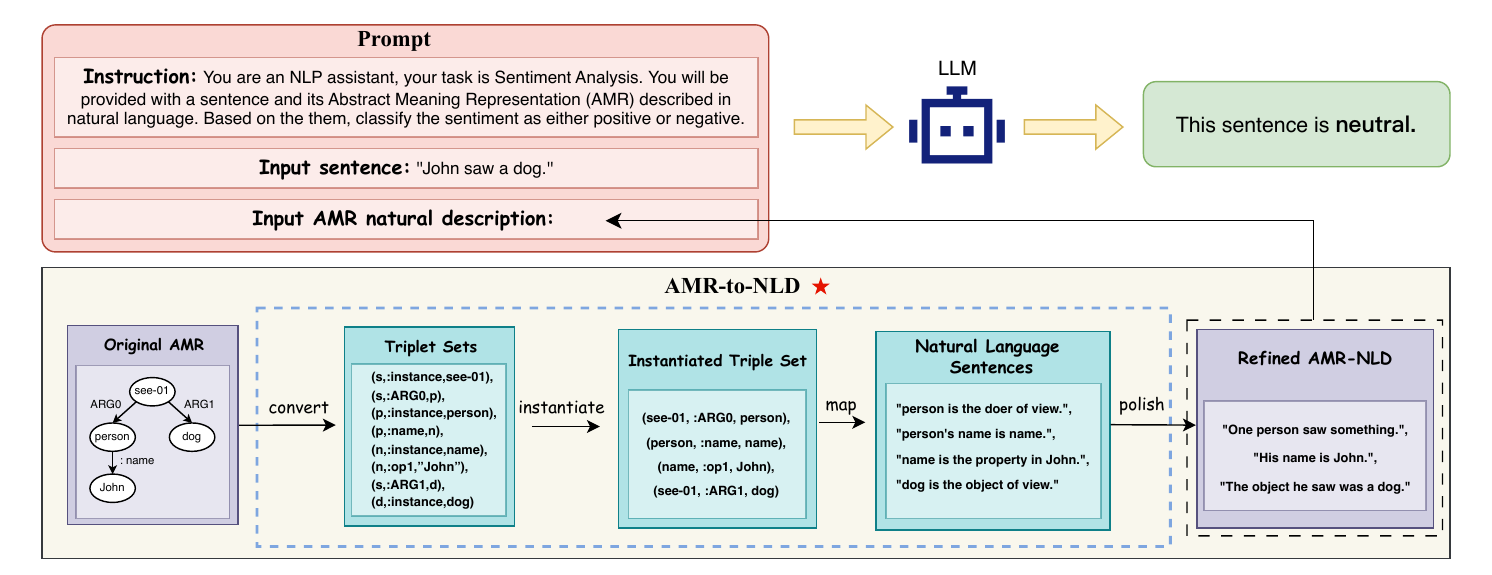}
\vspace{-0.3cm}
\caption{\textbf{The whole process of SR-LLM in training-free setting}. Initially, a task-specific prompt consists of an instruction, input sentence, and input SR structure (AMR is used here). Subsequently, the original AMR undergoes transformation via the \textbf{AMR-to-NLD} module, which employs predefined rules to map the AMR into an easily interpretable natural language description. This description is then subjected to refinement by a language model, ensuring fluency and coherence, resulting in \textbf{AMR-NLD}. Finally, the \textbf{AMR-NLD} is seamlessly integrated into the input, which is then fed into the LLM to generate the ultimate response.}

\label{fig:srllmtf}
\vspace{-0.05in}
\end{figure*}

\section{Method}

This chapter introduces the SR-LLM framework,  a novel paradigm designed to investigate the efficacious integration of SR into LLMs. The SR-LLM framework encompasses two configurations: training-free and training-dependent. These configurations are designed to amalgamate various types of SR through differentiated methodologies, thereby enhancing the LLMs' capability to comprehend and exploit structured information.

\subsection{SR-LLM Training-Free}
\begin{figure}[h!]
\centering
\vspace{-0.2cm}
\includegraphics[width=1\linewidth]{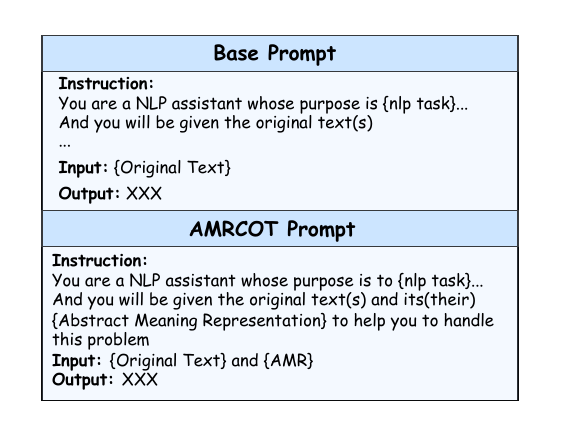}
\caption{\textbf{Base prompt and AMRCOT prompt.} \textbf{(Top)} This is the original task prompt, with only the raw text as input, serving as the standards for performance. \textbf{(Bottom)} This is the AMRCOT prompt method proposed by \citet{jin2024analyzing}, serving as a baseline.}
\label{fig:base_sr_cot}
\vspace{-0.2cm}
\end{figure}

Prior approaches, exemplified by AMRCOT \cite{jin2024analyzing}, have attempted to explicitly incorporate AMR into Chain-of-Thought (COT) prompts, as illustrated in Figure~\ref{fig:base_sr_cot}, have shown that this explicit approach fails to yield performance enhancement. We hypothesize that one factor contributing to this ineffectiveness stems from the inherent difficulty LLMs face in adequately comprehending and processing abstract structures such as AMR. In view of the aforementioned challenge, as illustrated in Figure~\ref{fig:srllmtf}, we propose SR-LLM Training-Free, where the original structured representation $Z$ is transformed into natural language descriptions termed \textbf{SR-NLD}, where SR can be instantiated with specific structured representations such as AMR, PST, and FOL. We refer to this entire transformation process as \textbf{SR-to-NLD}(\textbf{S}tructured \textbf{R}epresentation \textbf{to} \textbf{N}atural \textbf{L}anguage \textbf{D}escription). Specifically, the structured representations are mapped through predefined transformation rules, converting abstract symbols into easily interpretable natural language expressions. These generated natural language descriptions are then refined by a language model to ensure fluency and coherence. Finally, these descriptions are incorporated into the prompt and input into the target LLM. A pivotal advantage of this methodology lies in its training-free nature, as it does not require any additional fine-tuning or retraining of the LLM. Consequently, this technique offers remarkable flexibility, enabling rapid adaption to a diverse array of NLP tasks. 

\begin{algorithm}
\normalsize
    \caption{AMR-to-NLD Transformation}
    \label{algorithm_amr_nld}
    \begin{algorithmic}[1]
        \State \textbf{Input:} AMR graph $G = (V, E)$, nodes collection $V$, edges collection $E$, Penman library $\mathcal{P}$, language model $\theta$
        \State \textbf{Output:} Refined natural language descriptions $S_{\text{refined}}$
        
        \State \emph{\textbf{Phase 0}: Convert AMR to Triplets}
            \State Convert AMR graph $G$ into triplets $T = \{(c_1, r, c_2) \mid c_1, c_2 \in V, r \in E\}$ using the Penman library: $T=\mathcal{P}(G)$
        
        \State \emph{\textbf{Phase 1}: Identifier Instantiation}
            \For{each triplet $(c_1, r, c_2) \in T$}
                \If{$r = \texttt{:instance}$}
                    \State Replace identifiers $c_1, c_2$ with their corresponding concepts or instances
                \EndIf
            \EndFor

        \State \emph{\textbf{Phase 2}: Mapping to Natural Language}
            \State Convert triplets into natural language descriptions using a predefined dictionary: $M: T' \rightarrow S$
        
        \State \emph{\textbf{Phase 3}: Refinement}
            \State Refine the generated descriptions $S$ using language model: $S_{\text{refined}} = \theta (S)$
        
        \State \Return $S_{\text{refined}}$
    \end{algorithmic}
\end{algorithm}

Next, we shall elucidate the SR-to-NLD process, employing AMR-NLD as our quintessential exemplar, which shown in the Algorithm~\ref{algorithm_amr_nld}. The process first converts the AMR graph into triplets, then replaces the identifiers with actual concepts. Next, the triplets are mapped into natural language descriptions using predefined rules, and finally, the descriptions are refined by GPT-4o Mini to produce coherent AMR-NLD. To mitigate the risk of hallucination, we implemented a voting mechanism based on multiple generations. This detailed analysis forms the core of our discussion, outlining each step of the conversion process. The transformation methods for other SRs are elaborated in the Appendix~\ref{app:Detail_sr2nld} for completeness. Different from traditional SR-to-Text approaches, which generate a structurally coherent and fluent text based on the SR, such as the ``input sentence'' in Figure~\ref{fig:base_sr_cot}. SR-to-NLD aims to collaboratively describe the structured information through multiple sentences, as illustrated by the Refined AMR-NLD in Figure~\ref{fig:base_sr_cot}.

\subsection{SR-LLM Training-Dependent}

\begin{figure}
\centering
\vspace{0in}
\includegraphics[width=1\linewidth]{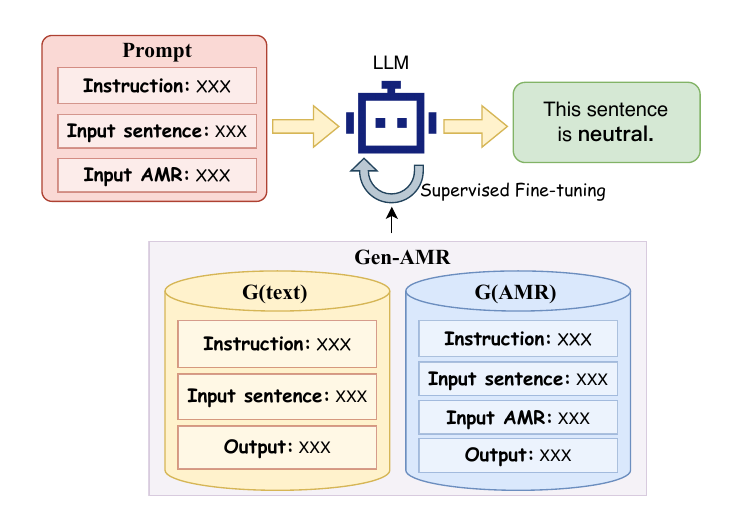}
\caption{\textbf{The whole process of SR-LLM in training-dependent setting}. Taking AMR as an example, a dataset called \textbf{Gen-AMR}, created by combining inputs consisting of sentences and their corresponding AMR structures, is utilized for the SFT of LLM to enhance the reasoning capability.}
\label{fig:srllmtd}
\vspace{-0.1in}
\end{figure}

In addition to making SRs more interpretable for LLMs, we also believe that establishing connections between tasks and structured information presents a potential opportunity. As shown in the Figure~\ref{fig:srllmtd}, in SR-LLM Training-Dependent, we constructed a task-specific hybrid dataset, named Gen-SR, where SR can be replaced by specific representations such as AMR, PST, and FOL. 

The entire hybrid dataset is composed of two parts: one consists of task-specific instruction pairs based on original text, while the other adds SRs in the instruction pairs based on the former. The former we mark as G(text) and the other we mark as G(SR). The complete example of these two are shown in the Appendix~\ref{app:prompt_gensr}. This mixed approach allows LLM to not only learn instruction-following for downstream tasks from G(text), but also to establish more robust connections between tasks and structures from G(SR), making the model achieve more effective improvements compared to learning solely from text.






\section{Experiments}
\label{experiment}

\subsection{Datasets}
To ensure comprehensive and diverse experiments, we selected 10 datasets covering various NLP tasks based on~\citet{liu2024datasets}, including five tasks from~\citet{jin2024analyzing} for result comparability. The dataset composition includes: PAWS for paraphrase detection~\citep{zhang2019paws}, SNLI for textual entailment recognition~\citep{bowman2015large}, WMT16 for translation tasks~\citep{bojar2016findings}, CoNLL2003 for named entity recognition~\citep{sang2003introduction}, Logic for logical fallacy detection~\citep{jin2022logical}, SST-2 for sentiment analysis~\citep{socher2013recursive}, Pubmed45 for event extraction~\citep{garg2016extracting}, WiC for word sense disambiguation~\citep{pilehvar2018wic}, SPIDER for Text2SQL code generation~\citep{yu2018spider}, and AGNEWS for text classification~\citep{zhang2015character}. \\
Regarding the source of SR datasets, we used a dual-source strategy: one part includes high-quality AMR datasets from Jin~\cite{jin2024analyzing}, covering five core tasks; the other is automatically constructed using GPT-4o, comprising supplementary AMR, PST, and FOL data. The detailed collection processes and results provided in the Appendix~\ref{app:dataset_tf}.

\subsection{Training-Free Results}

\setlength{\tabcolsep}{5.5pt}
\renewcommand{\arraystretch}{1.2}
\begin{table*}
\vspace{-0.2cm}
\caption{\textbf{Performance of SR-LLM(training-free). } In the table, a checkmark under ``SR'' indicates that the original SR was added to the prompt, while a checkmark under ``SR-NLD'' (highlighted with a gray background) represents the inclusion of SR-NLD in the prompt, which corresponds to the results of SR-LLM (training-free). No checkmarks indicate the use of the original prompt, serving as the control group for comparison. Our focus is on the performance differences between adding SR and SR-NLD, as well as their respective differences compared to the control group.
}
\vspace{-0.1cm}
\label{tab:tfperfor}
\footnotesize
\begin{tabular}{c|c|cccccccccc}
\toprule
SR & \begin{tabular}[c]{@{}c@{}}SR-NLD\\ (Ours)\end{tabular} & \begin{tabular}[c]{@{}c@{}}PAWS\\ (F1)\end{tabular} & \begin{tabular}[c]{@{}c@{}}Logic\\ (F1)\end{tabular} & \begin{tabular}[c]{@{}c@{}}Pubmed45\\ (F1)\end{tabular} & \begin{tabular}[c]{@{}c@{}}AGNEWS\\ (F1)\end{tabular} & \begin{tabular}[c]{@{}c@{}}WiC\\ (F1)\end{tabular} & \begin{tabular}[c]{@{}c@{}}SNLI\\ (F1)\end{tabular} & \begin{tabular}[c]{@{}c@{}}CoNLL\\2003\\(F1)\end{tabular} & \begin{tabular}[c]{@{}c@{}}SST-2\\ (F1)\end{tabular} & \multicolumn{1}{c}{\begin{tabular}[c]{@{}c@{}}WMT16\\ (BLEU)\end{tabular}} & \multicolumn{1}{c}{\begin{tabular}[c]{@{}c@{}}SPIDER\\ (F1)\end{tabular}} \\
\midrule
\multicolumn{12}{c}{(a) Llama3.1-
8b-Instruct} \\
\multicolumn{1}{l}{} &  & 41.59 & 15.48 & 24.35 & 53.88 & 43.99 & 25.81 & 46.28 & 68.72 & 13.16 & 24.80 \\
\checkmark &  & 36.41 & 14.20 & 20.69 & 48.17 & 42.05 & 23.17 & 41.75 & 65.66 & 12.34 & 21.53 \\
 \rowcolor{gray!15} 
 & \checkmark & \textbf{44.77} & \textbf{18.27} & \textbf{26.10} & \textbf{56.67} & \textbf{48.17} & \textbf{28.87} & \textbf{48.73} & \textbf{71.77} & \textbf{14.10} & \textbf{29.60} \\
\midrule
\multicolumn{12}{c}{(b) GPT 3.5-turbo} \\
 &  & 56.94 & 38.63 & 27.14 & \textbf{85.12} & 50.61 & 38.93 & \textbf{56.52} & 90.46 & 26.13 & 39.63 \\
\checkmark &  & 56.10 & 36.27 & 25.63 & 81.33 & 51.60 & 32.00 & 54.67 & 86.90 & 25.77 & 39.07 \\
\rowcolor{gray!15} 
 & \checkmark & \textbf{57.97} & \textbf{39.40} & \textbf{28.17} & 84.07 & \textbf{55.27} & \textbf{41.47} & 55.17 & \textbf{92.60} & \textbf{27.07} & \textbf{42.27} \\
\midrule
\multicolumn{12}{c}{(c) GPT 4o-mini} \\
 &  & 75.80 & \textbf{48.10} & \textbf{38.65} & \textbf{85.26} & \textbf{58.47} & 40.59 & \textbf{65.27} & 91.39 & \textbf{26.80} & 41.55 \\
\checkmark &  & 73.50 & 47.32 & 33.11 & 81.62 & 46.65 & 41.30 & 59.21 & 91.01 & 26.21 & 39.33 \\
\rowcolor{gray!15} 
 & \checkmark & \textbf{76.48} & 47.95 & 36.66 & 83.45 & 56.63 & \textbf{42.00} & 64.12 & \textbf{92.83} & 26.76 & \textbf{43.57}\\
\bottomrule
\end{tabular}
\vspace{-0.3cm}
\end{table*}

\paragraph{Experimental Details.}
We conducted experiments on the Llama3.1-8b-Instruct~\cite{dubey2024llama}, GPT-3.5-turbo, and GPT-4o-mini~\cite{achiam2023gpt} models, arranged from weak to strong according to their performance levels, employing two prompting strategies: Chain-of-Thought (CoT)~\cite{wei2022chain} and One-Shot~\cite{brown2020language}. CoT guides step-by-step reasoning, while One-Shot demonstrates task-solving through specific examples. All experiments were conducted independently on three types of SRs: AMR, FOL, and PST. Both PST and FOL were incorporated into the prompts using the same approach as AMRCOT~\cite{jin2024analyzing}. For brevity, the results obtained from these experiments were averaged and presented. For detailed prompts, refer to the Appendix~\ref{app:prompt}.

\paragraph{Result Analysis.} First, as shown in Table~\ref{tab:tfperfor}, incorporating SR-NLD into the prompt consistently outperforms incorporating the original SR. This indicates that for LLMs, transforming abstract SRs into natural language formats more familiar to the models is an effective strategy for enhancing their ability to interpret and apply structured information. Meanwhile, the comparision of the three models also reveals that the gradual decrease in the benefit of structured information as model performance increases. Specifically, for the Llama3.1-8b-Instruct model, results with SR-NLD significantly and consistently surpass those of the original prompt (i.e., without SR or SR-NLD). For GPT-3.5-turbo, most results show improvement, whereas for GPT-4o-mini, approximately half of the results demonstrate improvement, albeit with a smaller margin. This result further illustrates that weaker models benefit more from structured information as a supplement to the original text, aiding them in downstream reasoning tasks. In contrast, for stronger models, the additional structured information offers limited advantages and may even be less informative than the insights derived directly from the raw text.

\subsection{Training-Dependent Results}
\paragraph{Experimental Details}
We conducted experiments using the Llama3.1-8B-Instruct model to evaluate the performance of the training-dependent setting of SR-LLM, more detailed experimental parameters can be found in the Appendix~\ref{app:ftd}. The whole process of fine-tuning is a joint training across data from 10 tasks, rather than task-specific fine-tuning for any single dataset. Detailed data collection procedures and specific training data configurations are provided in the Appendix~\ref{app:dataset_td}. 
To provide a comparative analysis, we conducted three sets of experiments using the following datasets: 100\%G (text), 100\%G (SR), and a 50\%G (text) mixed with 50\% G (SR). The 50\%-50\% ratio was chosen because we considered it to be the most balanced approach. Further experiments, elaborated in Appendix~\ref{app:ratio_select}, also confirmed that this is the optimal mixing ratio. And we employed a random sampling approach. All experiments were conducted independently on three types of SRs and for brevity, the results obtained from these experiments were averaged and presented.

\setlength{\tabcolsep}{4.5pt}
\renewcommand{\arraystretch}{1.2}
\begin{table*}
\caption{\textbf{Performance of SR-LLM(training-dependent).} G(text) and G(SR) represent the types of training data, with 50\% and 10\% indicating their respective proportions in the total training dataset. Our focus is on the best performance of the model across various tasks under different fine-tuning strategies, as well as the performance differences between adding SR and the control group.
}
\vspace{-0.2cm}
\label{tab:tdperfor}
\footnotesize
\begin{tabular}{c|c|cccccccccc}
\toprule
FT Strategy & SR & \begin{tabular}[c]{@{}c@{}}PAWS\\ (F1)\end{tabular} & \begin{tabular}[c]{@{}c@{}}Logic\\ (F1)\end{tabular} & \begin{tabular}[c]{@{}c@{}}Pubmed45\\ (F1)\end{tabular} & \begin{tabular}[c]{@{}c@{}}AGNEWS\\ (F1)\end{tabular} & \begin{tabular}[c]{@{}c@{}}WiC\\ (F1)\end{tabular} & \begin{tabular}[c]{@{}c@{}}SNLI\\ (F1)\end{tabular} & \begin{tabular}[c]{@{}c@{}}CoNLL\\2003\\ (F1)\end{tabular} & \begin{tabular}[c]{@{}c@{}}SST-2\\ (F1)\end{tabular} & \begin{tabular}[c]{@{}c@{}}WMT16\\ (BLEU)\end{tabular} & \begin{tabular}[c]{@{}c@{}}SPIDER\\ (EM)\end{tabular} \\
 \midrule
\multirow{2}{*}{-} &  & 41.59 & 15.48 & 24.35 & 53.88 & 43.99 & 25.81 & 46.28 & 68.72 & 13.16 & 24.80 \\
 & \checkmark & 36.41 & 14.20 & 20.69 & 48.17 & 42.05 & 23.17 & 41.75 & 65.66 & 12.34 & 21.53 \\
  \midrule
\multirow{2}{*}{100\% G(text)} &  & 68.94 & 26.21 & 78.91 & 76.52 & 66.97 & 35.53 & 75.79 & 75.59 & 29.07 & 41.20 \\
 & \checkmark & 64.07 & 16.84 & 77.33 & 67.14 & 67.05 & 35.36 & 71.73 & 74.65 & 28.41 & 38.47 \\
  \midrule
\multirow{2}{*}{100\% G(SR)} &  & 65.34 & 25.23 & 81.13 & 75.10 & 66.44 & 36.68 & 75.40 & 77.49 & 26.93 & 37.07 \\
 & \checkmark & 75.39 & 29.89 & \textbf{82.02} & 81.99 & 70.82 & \textbf{56.62} & 76.27 & 81.62 & \textbf{30.80} & 40.60 \\
  \midrule
\multirow{2}{*}{\begin{tabular}[c]{@{}c@{}}50\% G(SR) \\ + 50\% G(text)\end{tabular}} &  & 68.66 & 26.77 & 79.78 & 75.77 & 69.48 & 36.49 & 75.42 & 77.13 & 26.14 & 42.40 \\
 & \checkmark & \textbf{81.04} & \textbf{36.52} & 81.85 & \textbf{82.63} & \textbf{74.68} & 54.92 & \textbf{76.67} & \textbf{83.72} & 30.33 & \textbf{48.93}\\
 \bottomrule
\end{tabular}
\vspace{-0.3cm}
\end{table*}

\paragraph{Result Analysis.}
As shown in the Table~\ref{tab:tdperfor}, when the fine-tuning dataset includes a certain proportion of SRs and incorporates SRs in the prompt, the model achieves superior performance in downstream tasks, consistently surpassing the case where the training data consists solely of text. Additionally, we observe that models fine-tuned with SRs data perform significantly better with prompts that include SRs, compared to the original prompts without SR. Conversely, when the training data consists entirely of text, the opposite trend is observed.
These findings suggest that when a model establishes a strong association between tasks and structured representations during training, it can leverage this information more effectively during inference. Furthermore, when the training data is entirely composed of structured representations, the performance is inferior to that achieved with a balanced mix of text and structured data. This highlights the critical importance of a balanced integration of raw text and structured representations in maximizing the model’s reasoning capabilities.

\subsection{Auxiliary Validation Experiments}

\paragraph{SR from High-Quality SR-Parsing Model.}
To validate the reliability of the generated SRs, we choose AMRBART~\cite{bai2022graph} to generate the required AMRs, and experiments were conducted to compare the results with those generated by GPT-4o. It is a model that demonstrates exceptional performance in the AMR parsing domain with a Smatch score of 85.4 on the AMR Parsing Leaderboard, ranking among the top-performing models.  As shown in the Table~\ref{tab:amrbart_per}, the performance differences between AMRs and AMR-NLDs derived from these two sources were minimal, almost always within 0.5\%. This indicates that the quality of the AMRs produced by AMRBART is comparable to those generated by our method. 

\setlength{\tabcolsep}{2.8pt}
\renewcommand{\arraystretch}{1.2}
\begin{table}
\caption{\textbf{Performance between different AMR Source. }Each data represents the performance difference of the model when using AMRs generated by GPT-4o versus AMRBART, calculated as the performance of AMRBART minus that of GPT-4o. As shown, the differences are almost all below 1\%.
}
\vspace{-0.1cm}
\label{tab:amrbart_per}
\footnotesize
\begin{tabular}{c|c|ccccc}
\toprule
AMR & \begin{tabular}[c]{@{}c@{}}AMR\\ (NLD)\end{tabular} & \begin{tabular}[c]{@{}c@{}}PAWS\\ (F1)\end{tabular} & \begin{tabular}[c]{@{}c@{}}Logic\\ (F1)\end{tabular} & \begin{tabular}[c]{@{}c@{}}Pubmed\\45\\ (F1)\end{tabular} & \begin{tabular}[c]{@{}c@{}}WMT\\16\\ (BLEU)\end{tabular} & \begin{tabular}[c]{@{}c@{}}SPIDER\\ (EM)\end{tabular} \\
\midrule
\multicolumn{7}{c}{(a) Llama3.1-8b-Instruct} \\
\checkmark &  & 0.40 & -0.07 & 0.01 & 0.13 & 0.28 \\
 & \checkmark & 0.77 & -0.13 & 0.50 & -0.02 & -0.01 \\
 \midrule
\multicolumn{7}{c}{(b) GPT 3.5-turbo} \\
\checkmark &  & 0.45 & 0.57 & -0.15 & 0.08 & 0.12 \\
 & \checkmark & 0.02 & -2.40 & 0.52 & 0.23 & 0.21 \\
 \midrule
\multicolumn{7}{c}{(c) GPT 4o-mini} \\
\checkmark &  & 0.08 & 0.07 & 0.53 & 0.49 & 0.02 \\
 & \checkmark & -0.11 & 0.61 & 0.61 & -0.13 & -0.13\\
 \bottomrule
\end{tabular}
\vspace{-0.6cm}
\end{table}

\setlength{\tabcolsep}{4.5pt}
\renewcommand{\arraystretch}{1.2}
\begin{table*}
\caption{\textbf{Performance between different AMR Quality}. The numbers in parentheses represent the performance differences between adding AMR or AMR-NLD and the control group. `Flawed' means the AMR is ambiguous or structurally flawed. `Gold' means the AMR is double checked by human and LLM.}
\vspace{-0.1cm}
\label{tab:goldamr}
\footnotesize
\begin{tabular}{c|c|c|ccccc}
\toprule
AMR Quality & AMR & AMR-NLD & PAWS (F1) & Logic (F1) & Pubmed45 (F1) & WMT16 (BLEU) & SPIDER (EM) \\
\midrule
 & \multicolumn{7}{c}{(a) Llama3.1-8b-Instruct} \\
- & - & - & 42.19 & 14.32 & 23.67 & 13.66 & 22.58 \\
Flawed & \checkmark &  & 34.5 (-7.69) & 11.52 (-2.8) & 19.41 (-4.26) & 11.07 (-2.59) & 18.26 (-4.32) \\
Gold & \checkmark &  & 42.48 (+0.29) & 14.7 (+0.38) & 23.43 (-0.24) & 14.65 (+0.99) & 22.93 (+0.35) \\
Flawed &  & \checkmark & 32.91 (-9.29) & 11.56 (-2.76) & 18.39 (-5.28) & 11.06 (-2.6) & 18.49 (-4.09) \\
Gold &  & \checkmark & \textbf{46.96 (+4.76)} & \textbf{18.98 (+4.66)} & \textbf{28.62 (+4.95)} & \textbf{19.13 (+5.47)} & \textbf{28.02 (+5.44)} \\
\midrule
 & \multicolumn{7}{c}{(b) GPT 3.5-turbo} \\
- & - & - & 56.04 & 43.79 & 28.29 & 26.01 & 40.28 \\
Flawed & \checkmark &  & 51.57 (-4.47) & 41.58 (-2.21) & 25.71 (-2.58) & 23.79 (-2.22) & 36.66 (-3.62) \\
Gold & \checkmark &  & 54.53 (-1.51) & 44.7 (+0.91) & 29.47 (+1.19) & 26.17 (+0.15) & 39.77 (-0.51) \\
Flawed &  & \checkmark & 51.33 (-4.71) & 39.79 (-4.01) & 26.9 (-1.38) & 24.37 (-1.64) & 36.74 (-3.54) \\
Gold &  & \checkmark & \textbf{56.78 (+0.74)} & \textbf{46.49 (+2.7)} & \textbf{32.0 (+3.71)} & \textbf{28.72 (+2.71)} & \textbf{44.81 (+4.53)} \\
\midrule
 & \multicolumn{7}{c}{(c) GPT 4o-mini} \\
- & - & - & 68.71 & 44.95 & 37.07 & 29.02 & 40.05 \\
Flawed & \checkmark &  & 65.63 (-3.08) & 42.74 (-2.2) & 35.42 (-1.66) & 27.31 (-1.71) & 37.84 (-2.21) \\
Gold & \checkmark &  & 70.04 (+1.33) & 45.9 (+0.96) & 35.36 (-1.71) & 29.62 (+0.6) & 41.47 (+1.42) \\
Flawed &  & \checkmark & 62.63 (-6.07) & 41.46 (-3.49) & 34.51 (-2.56) & 26.64 (-2.38) & 37.30 (-2.76) \\
Gold &  & \checkmark & \textbf{70.13 (+1.42)} & \textbf{46.18 (+1.23)} & \textbf{39.17 (+2.09)} & \textbf{30.14 (+1.12)} & \textbf{41.54 (+1.48)}\\
\bottomrule
\end{tabular}
\vspace{-0.3cm}
\end{table*}

\paragraph{Gold AMR vs Flawed AMR.}
Additionally, we selected 70 AMR samples (labeled as ``Flawed'') with ambiguities or structural flaws from each of the 10 datasets and refined them using a dual-process correction strategy that combined AMRBART-generated results with manual adjustments, producing high-quality AMRs (labeled ``Gold''). Results in Table~\ref{tab:goldamr} show that AMR quality significantly impacts model performance. Using flawed AMRs led to performance declines for both direct AMR and AMR-NLD representations, with a more pronounced drop for AMR-NLD. This indirectly validates AMR-NLD’s ability to enhance LLMs’ understanding of AMR structures. In contrast, with high-quality AMRs, AMR-NLD substantially improved model performance, while direct AMR usage showed limited gains. These results demonstrate that combining high-quality AMR-NLD is more effective in helping models comprehend structured information. This effect is particularly pronounced when the quality of the AMR is high, leading to substantial performance gains.

\paragraph{Fine-tuning Larger Model.}
To validate the robustness of the proposed method, we selected Llama3.1-70B-Instruct and conducted training-dependent experiments, whose details were consistent with those described for the Llama3.1-8B-Instruct model above, in five tasks shown in the Table~\ref{tab:tdper70}. The SR used in these experiments was AMR, with a 50\%-50\% ratio. We can see that, after fine-tuning, the model demonstrated improvements on all tasks, with corresponding values turning positive, more than half of which exceeded 5\%. These results further validate the effectiveness of Training-Dependent method on larger-scale models.

\setlength{\tabcolsep}{3.5pt}
\renewcommand{\arraystretch}{1.2}
\begin{table}
\caption{\textbf{Performance of SR-LLM(training-dependent) in Llama3.1-70b-Instruct.} The numbers in parentheses represent the performance differences between adding SR and the control group. Our focus is on the performance variations across different models with different prompts.
}
\vspace{-0.1cm}

\label{tab:tdper70}
\footnotesize
\begin{tabular}{c|ccccc}
\toprule
AMR & \begin{tabular}[c]{@{}c@{}}PAWS\\ (F1)\end{tabular} & \begin{tabular}[c]{@{}c@{}}Logic\\ (F1)\end{tabular} & \begin{tabular}[c]{@{}c@{}}Pubmed45\\ (F1)\end{tabular} & \begin{tabular}[c]{@{}c@{}}WMT16\\ (BLEU)\end{tabular} & \begin{tabular}[c]{@{}c@{}}SPIDER\\ (EM)\end{tabular} \\
\midrule
\multicolumn{6}{c}{(a) Vanilla} \\
 & 68.00 & 47.13 & 63.95 & 28.65 & 33.71 \\
\checkmark & \begin{tabular}[c]{@{}c@{}}60.28\\ (-7.73)\end{tabular} & \begin{tabular}[c]{@{}c@{}}43.08\\ (-4.04)\end{tabular} & \begin{tabular}[c]{@{}c@{}}48.82\\ (-15.13)\end{tabular} & \begin{tabular}[c]{@{}c@{}}27.91\\ (-0.73)\end{tabular} & \begin{tabular}[c]{@{}c@{}}29.20\\ (-4.51)\end{tabular} \\
\midrule
\multicolumn{6}{c}{(b) 50\% G(AMR) + 50\% G(text) } \\
 & 74.74 & 54.57 & 76.51 & 33.73 & 47.06 \\
 \rowcolor{gray!15}
\checkmark & \begin{tabular}[c]{@{}c@{}}\textbf{84.56}\\ (\textbf{+9.81})\end{tabular} & \begin{tabular}[c]{@{}c@{}}\textbf{58.96}\\ (\textbf{+4.39})\end{tabular} & \begin{tabular}[c]{@{}c@{}}\textbf{81.54}\\ (\textbf{+5.03})\end{tabular} & \begin{tabular}[c]{@{}c@{}}\textbf{37.00}\\ (\textbf{+3.27})\end{tabular} & \begin{tabular}[c]{@{}c@{}}\textbf{53.84}\\ (\textbf{+6.78})\end{tabular}\\
\bottomrule
\end{tabular}
\vspace{-0.5cm}
\end{table}

\paragraph{Experiment on Traceable LLM.}
Since NLP datasets have been public for years, their role in modern LLM development is unclear. We validate our approach through experiments on OLMo~\cite{OLMo}, where the training and validation data sources are clearly documented, in five tasks shown in the Table~\ref{tab:olmo}. The results show that when the prompt includes SR, its performance is lower than when SR is not included. However, when SR is transformed into SR-NLD, the performance improves significantly. For instance, in the PAWS, the performance increases from 61.52\% to 65.40\%. This demonstrates the robustness and generalizability of our approach.

\setlength{\tabcolsep}{2.6pt}
\renewcommand{\arraystretch}{1.2}

\begin{table}
\caption{\textbf{Performance in OLMo.} 
}

\vspace{-0.2cm}

\label{tab:olmo}
\footnotesize
\begin{tabular}{ccccccc}
\toprule
SR & \begin{tabular}[c]{@{}c@{}}SR\_NLD \\ (Ours)\end{tabular} & \begin{tabular}[c]{@{}c@{}}PAWS\\ (F1)\end{tabular} & \begin{tabular}[c]{@{}c@{}}Logic\\ (F1)\end{tabular} & \begin{tabular}[c]{@{}c@{}}Pubmed45\\ (F1)\end{tabular} & \begin{tabular}[c]{@{}c@{}}WMT16\\ (BLEU)\end{tabular} & \begin{tabular}[c]{@{}c@{}}SPIDER\\ (EM)\end{tabular} \\
\midrule
 &  & 61.52 & 22.96 & 60.27 & 9.67 & 20.44 \\
\checkmark &  & 57.73 & 17.97 & 57.89 & 11.23 & 19.79 \\
 \rowcolor{gray!15}
 & \checkmark & \textbf{65.40} & \textbf{24.89} & \textbf{65.19} & \textbf{12.44} & \textbf{22.20}\\
\bottomrule
\end{tabular}
\vspace{-0.2cm}
\end{table}

\section{Related Work}
\label{gen_inst}

\paragraph{Structure Representations.}
The SRs, including AMR, PST, and FOL, each unique advantages and applications in specific areas. AMR uses rooted, labeled graphs to abstract syntactic details, offering concise and semantically rich representations~\citep{banarescu2013abstract}. PST, based on Chomsky's generative grammar, employs hierarchical trees to represent sentence syntax and word dependencies~\citep{chomsky2014aspects}. FOL, as a symbolic logic system, defines objects, their relations, and properties, serving as a key tool in formal logic and reasoning~\citep{enderton2001mathematical, barwise1977introduction}.

\paragraph{Structure Representations Transformation.} 
The SR transformation has long been a critical area of research. Much of the existing work has focused on SR-to-Text approaches, which generate fluent text that aligns with the structure of the SR~\cite{song2018graph, ribeiro2021structural, wang2020amr}. Meanwhile, a method known as canonical expressions employs rule-based techniques to convert structures into standardized natural language representations, primarily to resolve ambiguities in non-standard sentences~\cite{shin2021constrained, roy2024benchclamp}. Its outputs are essentially normalized texts rather than comprehensive descriptions of the SR’s full structure. In contrast, our SR-to-NLD approach preserves the integrity of structured information while enhancing its interpretability through natural language descriptions of the structure.

\paragraph{Structured Representations used for NLP in LLM.} With the rise of LLM, studies like \citet{hahn2022formal} showed these sequence to sequence model's ability to generalize across formal domains, though challenges like low interpretability and hallucinations persist~\citet{de2023structuring}. Integrating structured representations into LLMs has improved accuracy and interpretability. \citet{yao2024semantic} and \cite{shi2024compressing} combined AMR with LLMs for tasks like sentence simplification and Retrieval-Augmented Generation. Additionally, \citet{hahn2022formal} and \cite{kalyanpur2024llm} advanced formal specification and logical reasoning in LLMs. And \citet{an2024rethinking} identified "magic prompts" that improve the performance of NLP tasks by solely focusing on semantic parsing, without the need to provide the actual parsing results. However, \citet{jin2024analyzing} argued that simplely add AMR into prompt might sometimes hinder performance in certain NLP tasks. 


\section{Conclusion}


SR-LLM demonstrates significant progress in enhancing LLMs' reasoning capabilities through structured representations. Our evaluation across diverse NLP tasks revealed SR's potential in generating novel implicit information. We established a framework for integrating SR into LLMs, from prompt engineering to fine-tuning, providing valuable insights into structured information incorporation. These advancements led to substantial improvements in both training-free and training-fependent settings, highlighting the effectiveness of integrating semantic, syntactic, and logical features. As we refine SR-LLM, we anticipate further progress towards more interpretable, accurate, and versatile language models with enhanced reasoning capabilities in various applications.

\section{limitations}


Despite SR-NLD's promising performance in certain tasks, its effectiveness remains inconsistent across different LLMs. The rule-based conversion method may constrain flexibility. Future research should focus on developing a more robust and adaptive structured representation, exploring task-specific optimizations, and investigating advanced conversion techniques and novel model architectures. Expanding evaluation to diverse language models and datasets will be crucial to enhance the method's consistency, flexibility, and applicability in various NLP domains.

\bibliography{reference}

\begin{thebibliography}{45}
\providecommand{\natexlab}[1]{#1}

\bibitem[{Achiam et~al.(2023)Achiam, Adler, Agarwal, Ahmad, Akkaya, Aleman, Almeida, Altenschmidt, Altman, Anadkat et~al.}]{achiam2023gpt}
Josh Achiam, Steven Adler, Sandhini Agarwal, Lama Ahmad, Ilge Akkaya, Florencia~Leoni Aleman, Diogo Almeida, Janko Altenschmidt, Sam Altman, Shyamal Anadkat, et~al. 2023.
\newblock Gpt-4 technical report.
\newblock \emph{arXiv preprint arXiv:2303.08774}.

\bibitem[{An et~al.(2024)An, Si, Hu, Zhao, Wang, Guo, and Chang}]{an2024rethinking}
Kaikai An, Shuzheng Si, Helan Hu, Haozhe Zhao, Yuchi Wang, Qingyan Guo, and Baobao Chang. 2024.
\newblock Rethinking semantic parsing for large language models: Enhancing llm performance with semantic hints.
\newblock \emph{arXiv preprint arXiv:2409.14469}.

\bibitem[{Bahdanau(2014)}]{bahdanau2014neural}
Dzmitry Bahdanau. 2014.
\newblock Neural machine translation by jointly learning to align and translate.
\newblock \emph{arXiv preprint arXiv:1409.0473}.

\bibitem[{Bai et~al.(2022)Bai, Chen, and Zhang}]{bai2022graph}
Xuefeng Bai, Yulong Chen, and Yue Zhang. 2022.
\newblock Graph pre-training for amr parsing and generation.
\newblock \emph{arXiv preprint arXiv:2203.07836}.

\bibitem[{Banarescu et~al.(2013)Banarescu, Bonial, Cai, Georgescu, Griffitt, Hermjakob, Knight, Koehn, Palmer, and Schneider}]{banarescu2013abstract}
Laura Banarescu, Claire Bonial, Shu Cai, Madalina Georgescu, Kira Griffitt, Ulf Hermjakob, Kevin Knight, Philipp Koehn, Martha Palmer, and Nathan Schneider. 2013.
\newblock Abstract meaning representation for sembanking.
\newblock In \emph{Proceedings of the 7th linguistic annotation workshop and interoperability with discourse}, pages 178--186.

\bibitem[{Barwise(1977)}]{barwise1977introduction}
Jon Barwise. 1977.
\newblock An introduction to first-order logic.
\newblock In \emph{Studies in Logic and the Foundations of Mathematics}, volume~90, pages 5--46. Elsevier.

\bibitem[{Bojar et~al.(2016)Bojar, Chatterjee, Federmann, Graham, Haddow, Huck, Yepes, Koehn, Logacheva, Monz et~al.}]{bojar2016findings}
Ondrej Bojar, Rajen Chatterjee, Christian Federmann, Yvette Graham, Barry Haddow, Matthias Huck, Antonio~Jimeno Yepes, Philipp Koehn, Varvara Logacheva, Christof Monz, et~al. 2016.
\newblock Findings of the 2016 conference on machine translation (wmt16).
\newblock In \emph{First conference on machine translation}, pages 131--198. Association for Computational Linguistics.

\bibitem[{Bowman et~al.(2015)Bowman, Angeli, Potts, and Manning}]{bowman2015large}
Samuel~R Bowman, Gabor Angeli, Christopher Potts, and Christopher~D Manning. 2015.
\newblock A large annotated corpus for learning natural language inference.
\newblock \emph{arXiv preprint arXiv:1508.05326}.

\bibitem[{Brown(2020)}]{brown2020language}
Tom~B Brown. 2020.
\newblock Language models are few-shot learners.
\newblock \emph{arXiv preprint arXiv:2005.14165}.

\bibitem[{Chomsky(2014)}]{chomsky2014aspects}
Noam Chomsky. 2014.
\newblock \emph{Aspects of the Theory of Syntax}.
\newblock 11. MIT press.

\bibitem[{Collobert et~al.(2011)Collobert, Weston, Bottou, Karlen, Kavukcuoglu, and Kuksa}]{collobert2011natural}
Ronan Collobert, Jason Weston, L{\'e}on Bottou, Michael Karlen, Koray Kavukcuoglu, and Pavel Kuksa. 2011.
\newblock Natural language processing (almost) from scratch.
\newblock \emph{Journal of machine learning research}, 12:2493--2537.

\bibitem[{Dagan et~al.(2005)Dagan, Glickman, and Magnini}]{dagan2005pascal}
Ido Dagan, Oren Glickman, and Bernardo Magnini. 2005.
\newblock The pascal recognising textual entailment challenge.
\newblock In \emph{Machine learning challenges workshop}, pages 177--190. Springer.

\bibitem[{Damonte et~al.(2016)Damonte, Cohen, and Satta}]{damonte2016incremental}
Marco Damonte, Shay~B Cohen, and Giorgio Satta. 2016.
\newblock An incremental parser for abstract meaning representation.
\newblock \emph{arXiv preprint arXiv:1608.06111}.

\bibitem[{De~Bellis(2023)}]{de2023structuring}
Alessandro De~Bellis. 2023.
\newblock Structuring the unstructured: an llm-guided transition.
\newblock In \emph{DC@ ISWC}.

\bibitem[{Dolan and Brockett(2005)}]{dolan2005automatically}
Bill Dolan and Chris Brockett. 2005.
\newblock Automatically constructing a corpus of sentential paraphrases.
\newblock In \emph{Third international workshop on paraphrasing (IWP2005)}.

\bibitem[{Dubey et~al.(2024)Dubey, Jauhri, Pandey, Kadian, Al-Dahle, Letman, Mathur, Schelten, Yang, Fan et~al.}]{dubey2024llama}
Abhimanyu Dubey, Abhinav Jauhri, Abhinav Pandey, Abhishek Kadian, Ahmad Al-Dahle, Aiesha Letman, Akhil Mathur, Alan Schelten, Amy Yang, Angela Fan, et~al. 2024.
\newblock The llama 3 herd of models.
\newblock \emph{arXiv preprint arXiv:2407.21783}.

\bibitem[{Enderton(2001)}]{enderton2001mathematical}
Herbert~B Enderton. 2001.
\newblock \emph{A mathematical introduction to logic}.
\newblock Elsevier.

\bibitem[{Garg et~al.(2016)Garg, Galstyan, Hermjakob, and Marcu}]{garg2016extracting}
Sahil Garg, Aram Galstyan, Ulf Hermjakob, and Daniel Marcu. 2016.
\newblock Extracting biomolecular interactions using semantic parsing of biomedical text.
\newblock In \emph{Proceedings of the AAAI Conference on Artificial Intelligence}, volume~30.

\bibitem[{Groeneveld et~al.(2024)Groeneveld, Beltagy, Walsh, Bhagia, Kinney, Tafjord, Jha, Ivison, Magnusson, Wang, Arora, Atkinson, Authur, Chandu, Cohan, Dumas, Elazar, Gu, Hessel, Khot, Merrill, Morrison, Muennighoff, Naik, Nam, Peters, Pyatkin, Ravichander, Schwenk, Shah, Smith, Strubell, Subramani, Wortsman, Dasigi, Lambert, Richardson, Zettlemoyer, Dodge, Lo, Soldaini, Smith, and Hajishirzi}]{OLMo}
Dirk Groeneveld, Iz~Beltagy, Pete Walsh, Akshita Bhagia, Rodney Kinney, Oyvind Tafjord, A.~Jha, Hamish Ivison, Ian Magnusson, Yizhong Wang, Shane Arora, David Atkinson, Russell Authur, Khyathi~Raghavi Chandu, Arman Cohan, Jennifer Dumas, Yanai Elazar, Yuling Gu, Jack Hessel, Tushar Khot, William Merrill, Jacob~Daniel Morrison, Niklas Muennighoff, Aakanksha Naik, Crystal Nam, Matthew~E. Peters, Valentina Pyatkin, Abhilasha Ravichander, Dustin Schwenk, Saurabh Shah, Will Smith, Emma Strubell, Nishant Subramani, Mitchell Wortsman, Pradeep Dasigi, Nathan Lambert, Kyle Richardson, Luke Zettlemoyer, Jesse Dodge, Kyle Lo, Luca Soldaini, Noah~A. Smith, and Hanna Hajishirzi. 2024.
\newblock \href {https://api.semanticscholar.org/CorpusID:267365485} {Olmo: Accelerating the science of language models}.
\newblock \emph{arXiv preprint}.

\bibitem[{Hahn et~al.(2022)Hahn, Schmitt, Tillman, Metzger, Siber, and Finkbeiner}]{hahn2022formal}
Christopher Hahn, Frederik Schmitt, Julia~J Tillman, Niklas Metzger, Julian Siber, and Bernd Finkbeiner. 2022.
\newblock Formal specifications from natural language.
\newblock \emph{arXiv preprint arXiv:2206.01962}.

\bibitem[{Jin et~al.(2024)Jin, Chen, Gonzalez, Liu, Zhang, Michael, Sch{\"o}lkopf, and Diab}]{jin2024analyzing}
Zhijing Jin, Yuen Chen, Fernando Gonzalez, Jiarui Liu, Jiayi Zhang, Julian Michael, Bernhard Sch{\"o}lkopf, and Mona Diab. 2024.
\newblock Analyzing the role of semantic representations in the era of large language models.
\newblock \emph{arXiv preprint arXiv:2405.01502}.

\bibitem[{Jin et~al.(2022)Jin, Lalwani, Vaidhya, Shen, Ding, Lyu, Sachan, Mihalcea, and Schoelkopf}]{jin2022logical}
Zhijing Jin, Abhinav Lalwani, Tejas Vaidhya, Xiaoyu Shen, Yiwen Ding, Zhiheng Lyu, Mrinmaya Sachan, Rada Mihalcea, and Bernhard Schoelkopf. 2022.
\newblock Logical fallacy detection.
\newblock \emph{arXiv preprint arXiv:2202.13758}.

\bibitem[{Johnson et~al.(2017)Johnson, Schuster, Le, Krikun, Wu, Chen, Thorat, Vi{\'e}gas, Wattenberg, Corrado et~al.}]{johnson2017google}
Melvin Johnson, Mike Schuster, Quoc~V Le, Maxim Krikun, Yonghui Wu, Zhifeng Chen, Nikhil Thorat, Fernanda Vi{\'e}gas, Martin Wattenberg, Greg Corrado, et~al. 2017.
\newblock Google’s multilingual neural machine translation system: Enabling zero-shot translation.
\newblock \emph{Transactions of the Association for Computational Linguistics}, 5:339--351.

\bibitem[{Kalyanpur et~al.(2024)Kalyanpur, Saravanakumar, Barres, Chu-Carroll, Melville, and Ferrucci}]{kalyanpur2024llm}
Aditya Kalyanpur, Kailash Saravanakumar, Victor Barres, Jennifer Chu-Carroll, David Melville, and David Ferrucci. 2024.
\newblock Llm-arc: Enhancing llms with an automated reasoning critic.
\newblock \emph{arXiv preprint arXiv:2406.17663}.

\bibitem[{Knight et~al.(2020)Knight, Badarau, Baranescu, Bonial, Bardocz, Griffitt, Hermjakob, Marcu, Palmer, O'Gorman et~al.}]{knight2020abstract}
Kevin Knight, Bianca Badarau, Laura Baranescu, Claire Bonial, Madalina Bardocz, Kira Griffitt, Ulf Hermjakob, Daniel Marcu, Martha Palmer, Tim O'Gorman, et~al. 2020.
\newblock \href {https://doi.org/10.35111/44cy-bp51} {Abstract meaning representation (amr) annotation release 3.0}.
\newblock Web Download.
\newblock LDC2020T02.

\bibitem[{Liu et~al.(2024)Liu, Cao, Liu, Ding, and Jin}]{liu2024datasets}
Yang Liu, Jiahuan Cao, Chongyu Liu, Kai Ding, and Lianwen Jin. 2024.
\newblock Datasets for large language models: A comprehensive survey.
\newblock \emph{arXiv preprint arXiv:2402.18041}.

\bibitem[{Manning(1999)}]{manning1999foundations}
Christopher~D Manning. 1999.
\newblock \emph{Foundations of statistical natural language processing}.
\newblock The MIT Press.

\bibitem[{Mihalcea et~al.(2006)Mihalcea, Corley, Strapparava et~al.}]{mihalcea2006corpus}
Rada Mihalcea, Courtney Corley, Carlo Strapparava, et~al. 2006.
\newblock Corpus-based and knowledge-based measures of text semantic similarity.
\newblock In \emph{Aaai}, 2006, pages 775--780.

\bibitem[{Pilehvar and Camacho-Collados(2018)}]{pilehvar2018wic}
Mohammad~Taher Pilehvar and Jose Camacho-Collados. 2018.
\newblock Wic: the word-in-context dataset for evaluating context-sensitive meaning representations.
\newblock \emph{arXiv preprint arXiv:1808.09121}.

\bibitem[{Ram{\'\i}rez(2024)}]{ramirez2024natural}
Jos{\'e} Gabriel~Carrasco Ram{\'\i}rez. 2024.
\newblock Natural language processing advancements: Breaking barriers in human-computer interaction.
\newblock \emph{Journal of Artificial Intelligence General Science (JAIGS) ISSN: 3006-4023}, 3(1):31--39.

\bibitem[{Ribeiro et~al.(2021)Ribeiro, Zhang, and Gurevych}]{ribeiro2021structural}
Leonardo~FR Ribeiro, Yue Zhang, and Iryna Gurevych. 2021.
\newblock Structural adapters in pretrained language models for amr-to-text generation.
\newblock \emph{arXiv preprint arXiv:2103.09120}.

\bibitem[{Roy et~al.(2024)Roy, Thomson, Chen, Shin, Pauls, Eisner, and Van~Durme}]{roy2024benchclamp}
Subhro Roy, Samuel Thomson, Tongfei Chen, Richard Shin, Adam Pauls, Jason Eisner, and Benjamin Van~Durme. 2024.
\newblock Benchclamp: A benchmark for evaluating language models on syntactic and semantic parsing.
\newblock \emph{Advances in Neural Information Processing Systems}, 36.

\bibitem[{Sachan et~al.(2020)Sachan, Zhang, Qi, and Hamilton}]{sachan2020syntax}
Devendra~Singh Sachan, Yuhao Zhang, Peng Qi, and William Hamilton. 2020.
\newblock Do syntax trees help pre-trained transformers extract information?
\newblock \emph{arXiv preprint arXiv:2008.09084}.

\bibitem[{Sang and De~Meulder(2003)}]{sang2003introduction}
Erik~F Sang and Fien De~Meulder. 2003.
\newblock Introduction to the conll-2003 shared task: Language-independent named entity recognition.
\newblock \emph{arXiv preprint cs/0306050}.

\bibitem[{Shi et~al.(2024)Shi, Sun, Li, and Xu}]{shi2024compressing}
Kaize Shi, Xueyao Sun, Qing Li, and Guandong Xu. 2024.
\newblock Compressing long context for enhancing rag with amr-based concept distillation.
\newblock \emph{arXiv preprint arXiv:2405.03085}.

\bibitem[{Shin et~al.(2021)Shin, Lin, Thomson, Chen, Roy, Platanios, Pauls, Klein, Eisner, and Van~Durme}]{shin2021constrained}
Richard Shin, Christopher~H Lin, Sam Thomson, Charles Chen, Subhro Roy, Emmanouil~Antonios Platanios, Adam Pauls, Dan Klein, Jason Eisner, and Benjamin Van~Durme. 2021.
\newblock Constrained language models yield few-shot semantic parsers.
\newblock \emph{arXiv preprint arXiv:2104.08768}.

\bibitem[{Socher et~al.(2013)Socher, Perelygin, Wu, Chuang, Manning, Ng, and Potts}]{socher2013recursive}
Richard Socher, Alex Perelygin, Jean Wu, Jason Chuang, Christopher~D Manning, Andrew~Y Ng, and Christopher Potts. 2013.
\newblock Recursive deep models for semantic compositionality over a sentiment treebank.
\newblock In \emph{Proceedings of the 2013 conference on empirical methods in natural language processing}, pages 1631--1642.

\bibitem[{Song et~al.(2018)Song, Zhang, Wang, and Gildea}]{song2018graph}
Linfeng Song, Yue Zhang, Zhiguo Wang, and Daniel Gildea. 2018.
\newblock A graph-to-sequence model for amr-to-text generation.
\newblock \emph{arXiv preprint arXiv:1805.02473}.

\bibitem[{Wang et~al.(2015)Wang, Xue, and Pradhan}]{wang2015transition}
Chuan Wang, Nianwen Xue, and Sameer Pradhan. 2015.
\newblock A transition-based algorithm for amr parsing.
\newblock In \emph{Proceedings of the 2015 Conference of the North American Chapter of the Association for Computational Linguistics: Human Language Technologies}, pages 366--375.

\bibitem[{Wang et~al.(2020)Wang, Wan, and Jin}]{wang2020amr}
Tianming Wang, Xiaojun Wan, and Hanqi Jin. 2020.
\newblock Amr-to-text generation with graph transformer.
\newblock \emph{Transactions of the Association for Computational Linguistics}, 8:19--33.

\bibitem[{Wei et~al.(2022)Wei, Wang, Schuurmans, Bosma, Xia, Chi, Le, Zhou et~al.}]{wei2022chain}
Jason Wei, Xuezhi Wang, Dale Schuurmans, Maarten Bosma, Fei Xia, Ed~Chi, Quoc~V Le, Denny Zhou, et~al. 2022.
\newblock Chain-of-thought prompting elicits reasoning in large language models.
\newblock \emph{Advances in neural information processing systems}, 35:24824--24837.

\bibitem[{Yao et~al.(2024)Yao, Guzhva, and Barbosa}]{yao2024semantic}
Peiran Yao, Kostyantyn Guzhva, and Denilson Barbosa. 2024.
\newblock Semantic graphs for syntactic simplification: A revisit from the age of llm.
\newblock \emph{arXiv preprint arXiv:2407.04067}.

\bibitem[{Yu et~al.(2018)Yu, Zhang, Yang, Yasunaga, Wang, Li, Ma, Li, Yao, Roman et~al.}]{yu2018spider}
Tao Yu, Rui Zhang, Kai Yang, Michihiro Yasunaga, Dongxu Wang, Zifan Li, James Ma, Irene Li, Qingning Yao, Shanelle Roman, et~al. 2018.
\newblock Spider: A large-scale human-labeled dataset for complex and cross-domain semantic parsing and text-to-sql task.
\newblock \emph{arXiv preprint arXiv:1809.08887}.

\bibitem[{Zhang et~al.(2015)Zhang, Zhao, and LeCun}]{zhang2015character}
Xiang Zhang, Junbo Zhao, and Yann LeCun. 2015.
\newblock Character-level convolutional networks for text classification.
\newblock \emph{Advances in neural information processing systems}, 28.

\bibitem[{Zhang et~al.(2019)Zhang, Baldridge, and He}]{zhang2019paws}
Yuan Zhang, Jason Baldridge, and Luheng He. 2019.
\newblock Paws: Paraphrase adversaries from word scrambling.
\newblock \emph{arXiv preprint arXiv:1904.01130}.

\end{thebibliography}

\bibliographystyle{acl_natbib}

\clearpage

\appendix
\section{Experimental Details}
\subsection{Details of Converting SR to SR-NLD}
\label{app:Detail_sr2nld}
\subsubsection{Details of Translating AMR Triplet to Natural Sentence} 

\begin{figure}[ht]
\centering
\vspace{0in}
\includegraphics[width=1\linewidth]{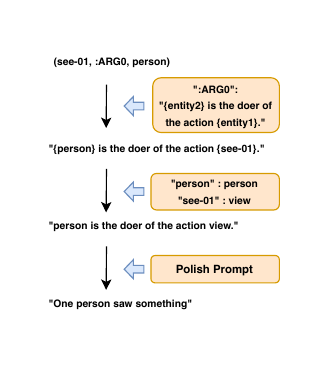}
\caption{The process of translate entities and relationships into natural language sentences}
\label{fig:detail_amr_nld}
\vspace{-0.1in}
\end{figure}

\begin{figure}[ht]
\centering
\vspace{0in}
\includegraphics[width=0.8\linewidth]{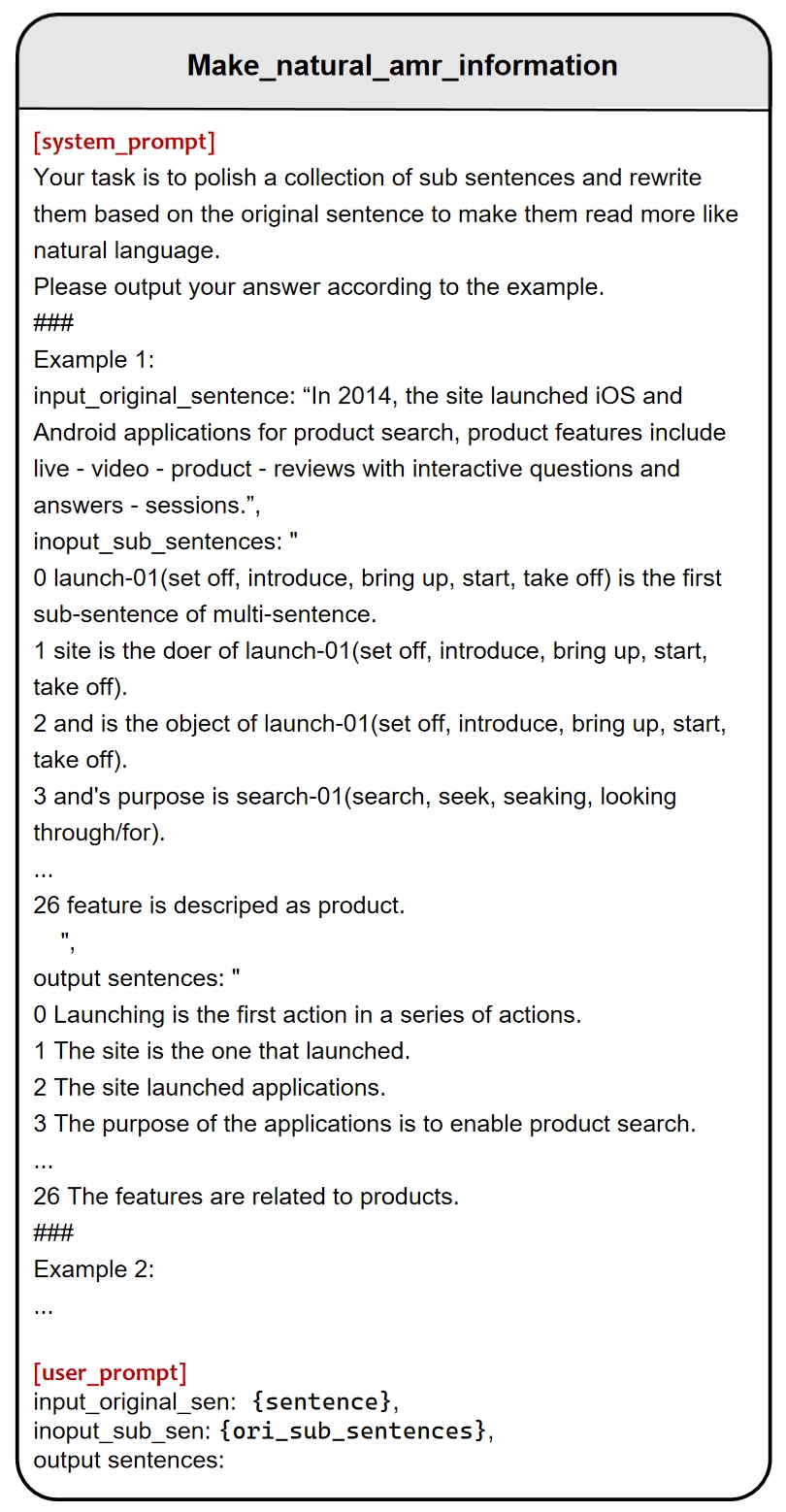}
\caption{The prompt of polishing sentence for making AMR-NLD}
\label{fig:amrnld_prompt}
\vspace{-0.1in}
\end{figure}

According to the Figure~\ref{fig:detail_amr_nld}, first, the triplet is converted into a sentence based on the relation mapping rules. Then, using the entity dictionary, the entities are replaced with their actual meanings to form the final sentence. Finally, the sentence is input into the LLM for refinement into a complete and coherent sentence, as shown in the Figure~\ref{fig:amrnld_prompt}.

\subsubsection{Whole Process of Making PST-NLD}
\label{app:make_pst_nld}
\begin{figure}[!ht]
\centering
\vspace{0.1in}
\includegraphics[width=1\linewidth]{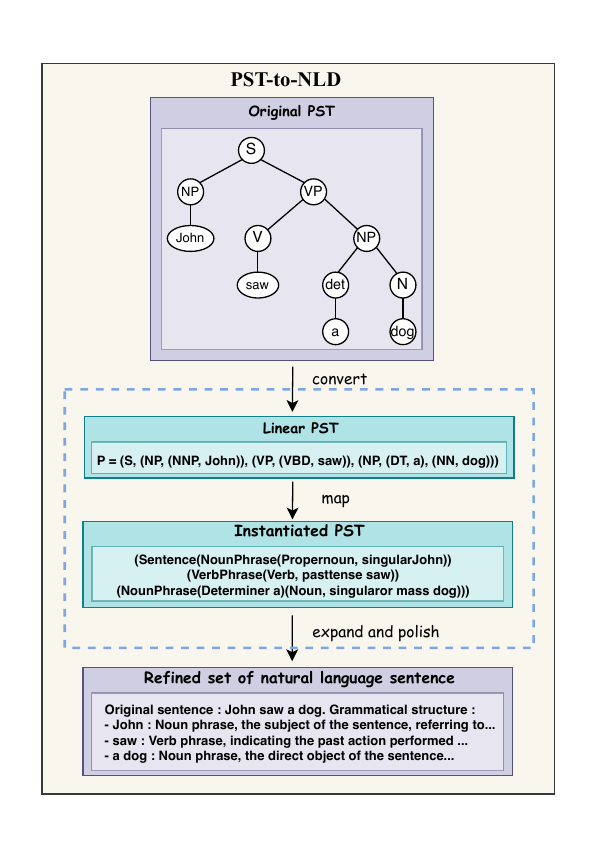}
\vspace{0.1in}
\caption{\textbf{The Whole process of Making PST-NLD.} The process of creating PST-NLD involves first converting the PST tree structure into a linear sequence of symbols using depth-first search (DFS). Then, a mapping function is applied to translate each node and its children into natural language descriptions. Finally, a language model is used to refine the generated descriptions, making them more natural and coherent.}
\label{fig:pst2nld}
\vspace{0.1in}
\end{figure}

\paragraph{Definition of PST.}
    PST is represented as a tree structure \( T = (N, E) \). Here \( N \) denotes the set of nodes, representing the syntactic components of a sentence (e.g., part-of-speech tags and phrase labels). Node types include \( S \) (sentence), \( NP \) (noun phrase), \( VP \) (verb phrase), etc.  \( E \) denotes the set of edges, representing dependencies between components.
    An example of the original PST structure is shown in the Figure~\ref{fig:pst2nld}.

\paragraph{Conversion of PST to a Linear Structure Using Depth-First Search (DFS).}
    Starting from the root node (typically \( n_0 \), representing the sentence's syntactic structure, such as \( S \)), we traverse the tree in a depth-first search (DFS) manner, converting it into a linear sequence of symbols \( P \).

\paragraph{Mapping PST Identifiers to Natural Language Descriptions.}

    We define a mapping function \( M \) to translate each identifier (e.g., \( S \), \( NP \), \( VBD \)) and its child nodes into natural language descriptions. The dictionary \( D \), which specifies the natural language interpretation of each identifier, is detailed in the appendix. For each triplet \( (n, c_1, c_2) \), where \( n \) is a node and \( c_1 \), \( c_2 \) are its children, we apply the mapping function \( M(n) = \text{description}(n) \). The resulting natural language description \( S \) is as shown in the Figure~\ref{fig:pst2nld}.
    
\paragraph{Refinement of Natural Language Descriptions Using a Language Model.}

\begin{figure}[ht]
\centering
\vspace{0in}
\includegraphics[width=0.8\linewidth]{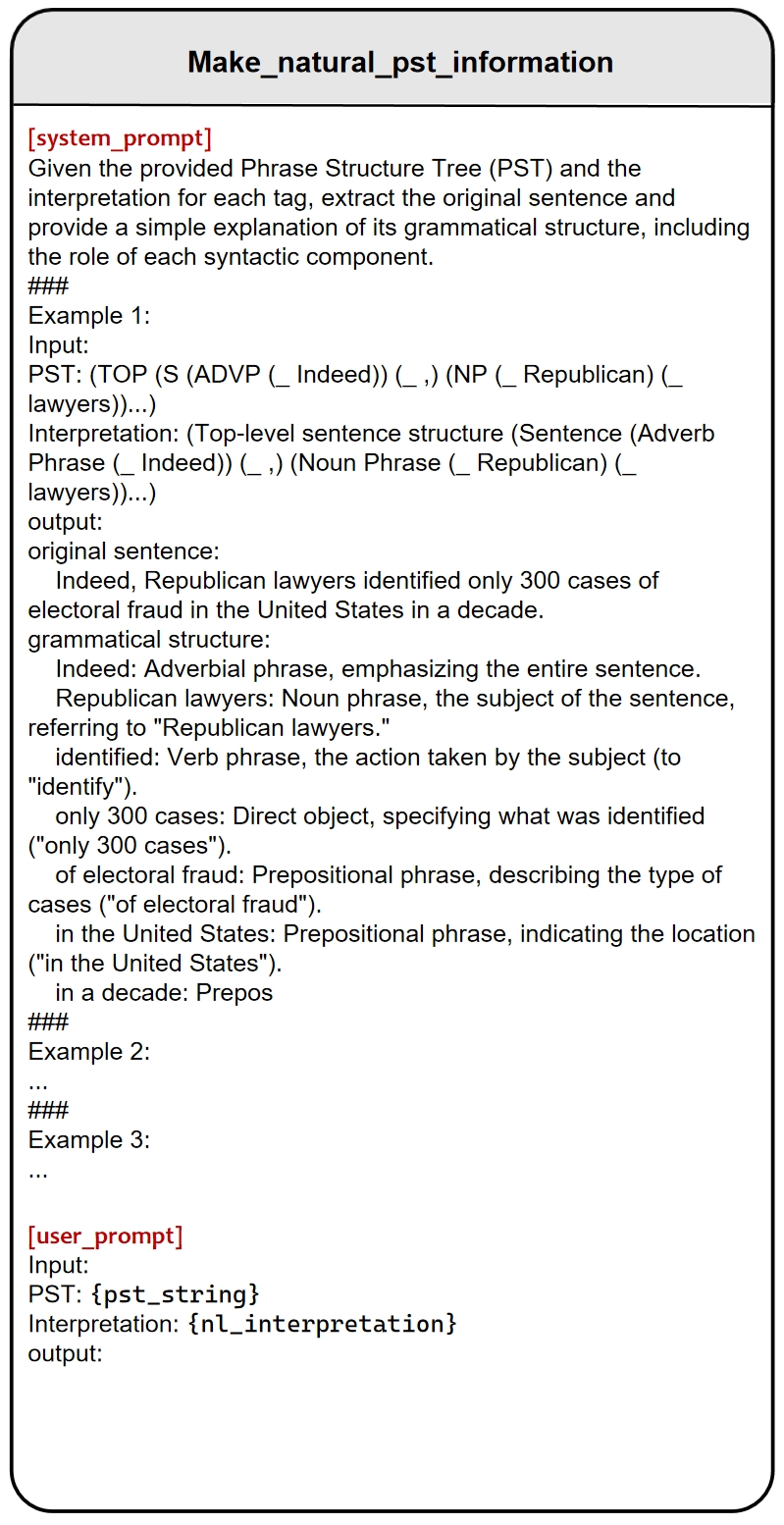}
\caption{The prompt of polishing sentence for making PST-NLD}
\label{fig:pstnld_prompt}
\vspace{-0.1in}
\end{figure}

    To make the descriptions more natural and coherent, the generated descriptions \( S \) are refined using the language model \( F_{\text{LM}}: S \rightarrow S_{\text{refined}} \). The specific prompt is shown in the prompt (b) of Figure~\ref{fig:pstnld_prompt}. 
  
\begin{figure}
\centering
\vspace{0.1in}
\includegraphics[width=1\linewidth]{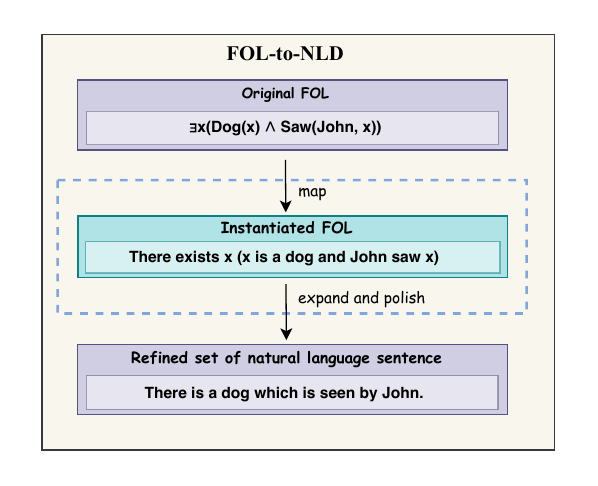}
\vspace{0.1in}
\caption{\textbf{The Whole process of Making FOL-NLD.} The process of converting FOL to NLD involves first mapping FOL symbols, such as variables, predicates, and logical operators, into natural language using predefined symbol mappings and logical rules. Then, the generated descriptions are refined using a language model to ensure they are coherent and fluent.}
\label{fig:fol2nld}
\vspace{0.1in}
\end{figure}

\subsubsection{Whole Process of Making FOL-NLD}
\label{app:make_fol_nld}
\paragraph{Definition of FOL.}
    FOL is represented as \( F = (Q, V, P, C) \), where 
         \( Q \) denotes the set of quantifiers, used to express the existence of variables, such as \( \exists \) (exists) and \( \forall \) (for all).
         \( V \) represents the set of variables, representing objects in FOL, typically denoted as \( x, y, z \).
         \( P \) represents the set of predicates, used to express properties of objects or relationships between multiple objects.
         \( C \) represents the set of logical connectives, used to connect multiple propositions, including conjunction (\( \land \)), disjunction (\( \lor \)), and negation (\( \neg \)).
    An example of the original FOL structure is shown in the Figure~\ref{fig:fol2nld}.

\paragraph{Mapping FOL to Natural Language Descriptions.}

We define a mapping function \( M = (D, L) \), where \( D \) is a set of symbol mappings that translates variables, predicates, and logical operators in FOL into natural language descriptions. \( L \) is a set of logical mapping rules that transforms the logical structure of FOL into natural language syntax. By applying these mapping rules to the initial FOL expressions, we can convert logical symbols into natural language descriptions.
    
\begin{figure}[ht]
\centering
\vspace{0in}
\includegraphics[width=0.8\linewidth]{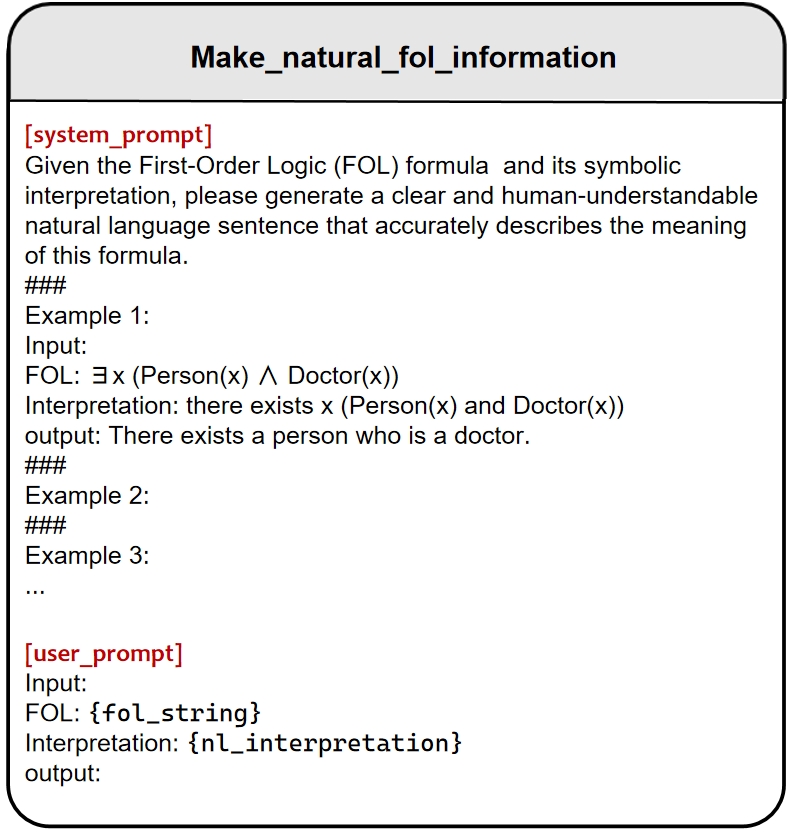}
\caption{The prompt of polishing sentence for making FOL-NLD}
\label{fig:folnld_prompt}
\vspace{-0.1in}
\end{figure}

\paragraph{Refinement of Natural Language Descriptions Using a Language Model.} To ensure that the descriptions are coherent and fluent, we refine the generated descriptions \( S \) using the language model \( F_{\text{LM}}: S \rightarrow S_{\text{refined}} \). The specific prompt is shown in the prompt (c) of Figure~\ref{fig:folnld_prompt}.

\subsection{Complete Fine-tuning Details}
\label{app:ftd}
We used Meta's Llama-3.1-8B-Instruct as the backbone and conducted fine-tuning on 8 NVIDIA A100-80G GPUs. Optimization was performed using the AdamW optimizer with a learning rate of 1e-4 and cosine learning rate decay. The training setup included a per\_device\_train\_batch\_size of 16 and gradient\_accumulation\_steps of 8, yielding an effective global batch size of 1024. A fixed random seed of 42 ensured reproducibility. Each dataset was fine-tuned for 10 epochs, with early stopping to prevent overfitting.

\section{Data Collection}
\subsection{The Process of Constructing Datasets for All Tasks of SR-LLM~(training-free)}
\label{app:dataset_tf}
In this section, I will outline the process of collecting test data for the 10 tasks used in SR-LLM (training-free), including both the original text and three types of structured representations. The data statistics are summarized in the Table~\ref{tab:dataset_of_tf}.

\setlength{\tabcolsep}{4.5pt}
\renewcommand{\arraystretch}{1.2}
\begin{table}[!ht]
\centering
\vspace{-0.1in}
\caption{Tasks and datasets used in SR-LLM (training-free)}
\label{tab:dataset_of_tf}
\vspace{-0.1in}
\small
\resizebox{0.5\textwidth}{!}
{ 
\begin{tabular}{lcc}
\toprule
\textbf{Dataset} & \textbf{Task} & \textbf{Test Size} \\
\midrule
PAWS      & Paraphrase Detection          & 8000  \\
SNLI      & Recognizing Textual Entailment & 10000 \\
WMT16     & Translation                   & 5999  \\
CoNLL2003 & Named Entity Recognition       & 3453  \\
LOGIC      & Logical Fallacy Detection     & 2449  \\
SST-2     & Sentiment Analysis            & 872   \\
Pubmed45  & Event Extraction              & 5000  \\
WiC       & Lexical Disambiguation        & 2038  \\
SPIDER    & Text2SQL Code Generation      & 8034  \\
AGNEWS    & Text Classification           & 7600  \\
\bottomrule
\end{tabular}
} 
\vspace{-0.1in}
\end{table}

\begin{figure}[ht]
\centering
\vspace{0in}
\includegraphics[width=0.8\linewidth]{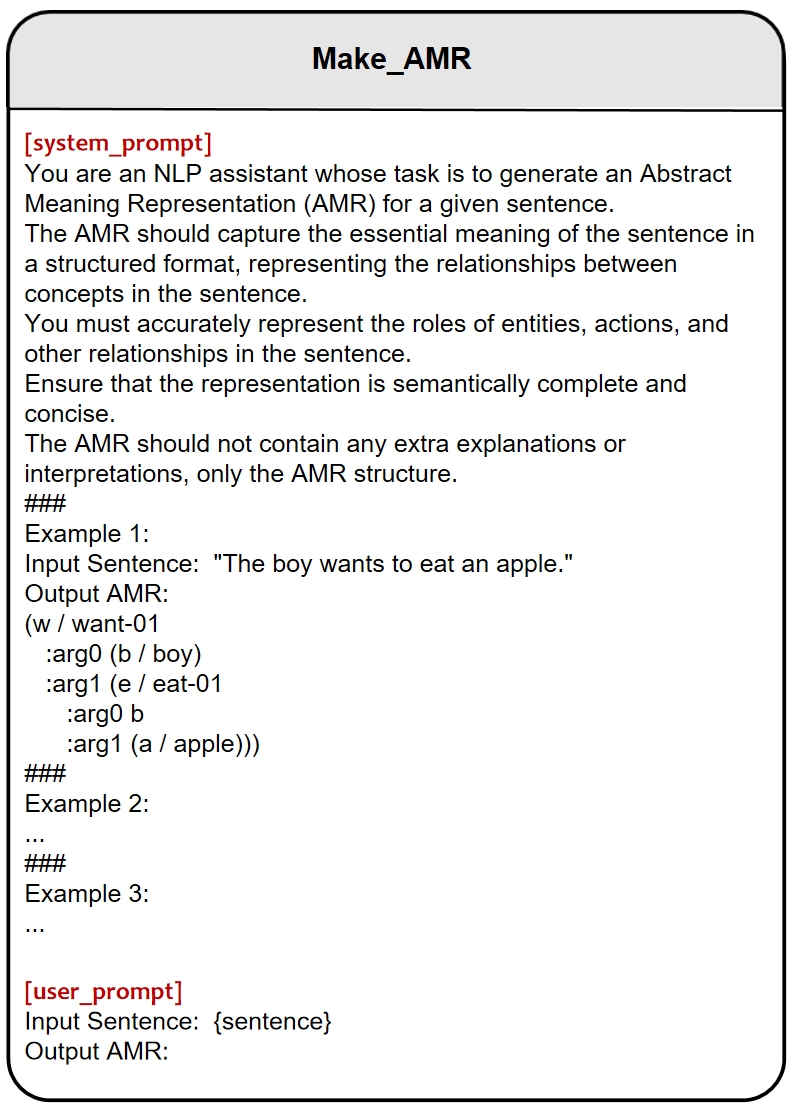}
\caption{The prompt of making AMR}
\label{fig:makr_amr}
\vspace{-0.1in}
\end{figure}

\begin{figure}[ht]
\centering
\vspace{0in}
\includegraphics[width=0.8\linewidth]{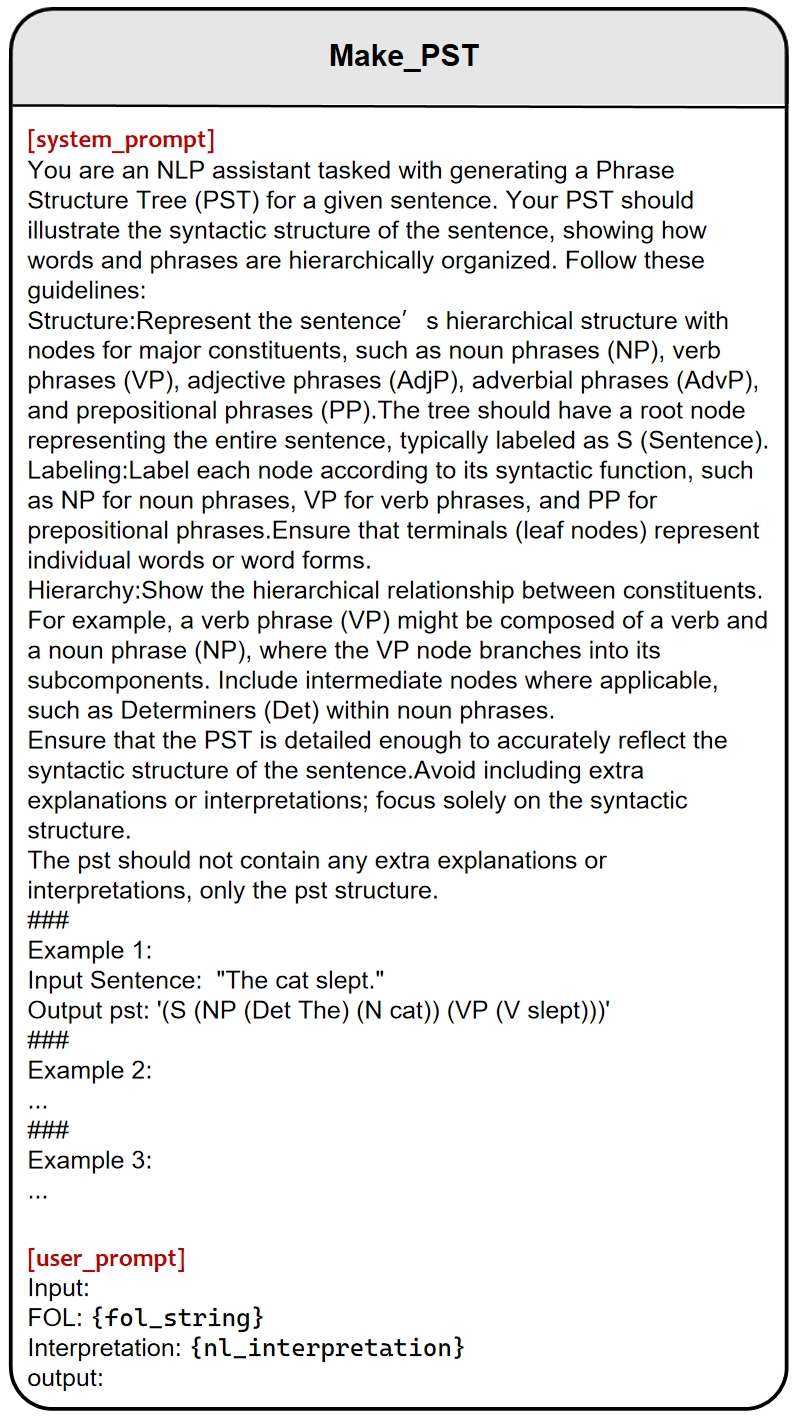}
\caption{The prompt of making PST}
\label{fig:makr_pst}
\vspace{-0.1in}
\end{figure}

\begin{figure}[ht]
\centering
\vspace{0in}
\includegraphics[width=0.8\linewidth]{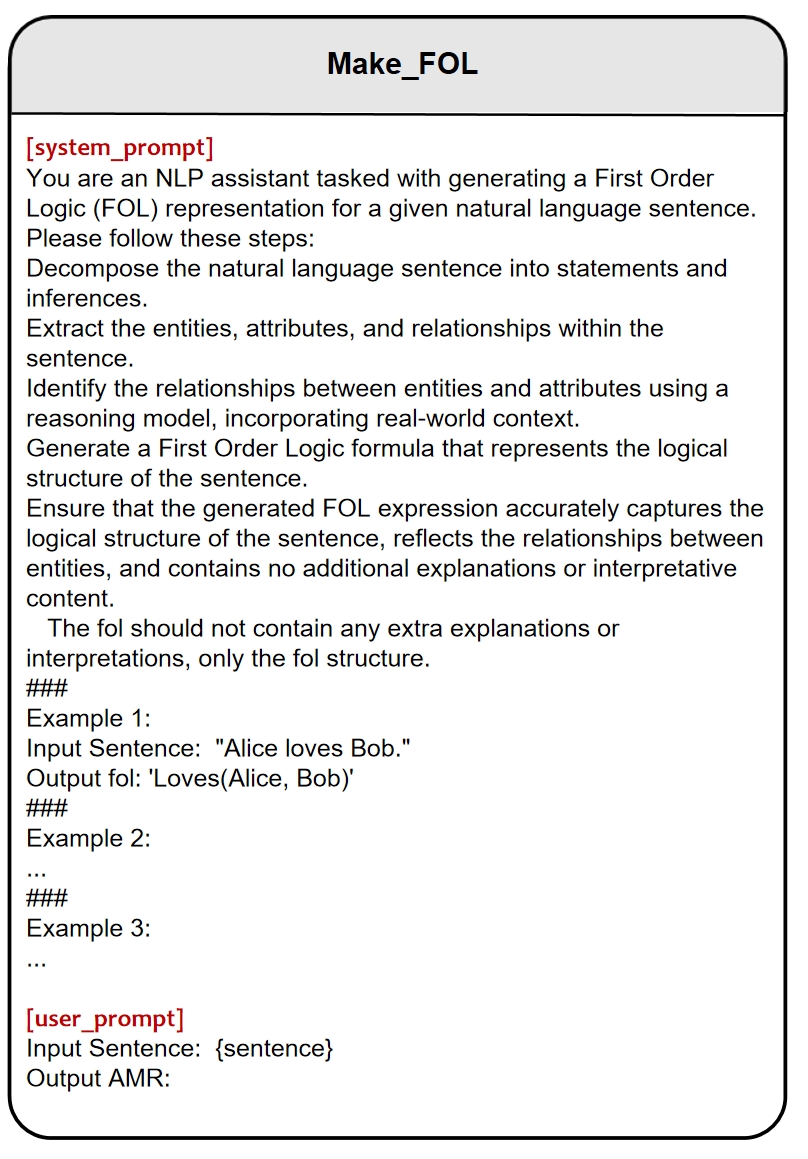}
\caption{The prompt of making FOL}
\label{fig:makr_fol}
\vspace{-0.1in}
\end{figure}

\paragraph{SNLI}
SNLI is a large and comprehensive dataset, with a test set containing 10,000 examples. Therefore, we directly used the test set for our experiments. The AMR, FOL, and PST data were generated using GPT-4o-turbo in a few-shots setting, with the prompt provided in the Figure~\ref{fig:makr_amr}, Figure~\ref{fig:makr_pst} and Figure~\ref{fig:makr_fol}.

\paragraph{CoNLL2003}
CoNLL2003 is also a rich and complete dataset, with a test set of 3,453 examples, which we used directly. Structured representations were generated using the same method as described above.

\paragraph{SST-2}
Since the official SST-2 test set does not contain labels, we used the full validation set of 872 examples as the test set for this experiment. Structured representations were generated using the same method as described above.

\paragraph{WiC}
The WiC test set consists of 1,400 examples, which is relatively small. Therefore, we combined the 648 examples from the validation set to create a larger test set. Structured representations were generated using the same method as described above.

\paragraph{AGNEWS}
AGNEWS is another large and comprehensive dataset, with a test set of 7,600 examples, which we used directly. Structured representations were generated using the same method as described above.

\paragraph{PAWS}
To ensure sufficient comparability in the experiments, the original text data and AMR representations for PAWS were sourced from \citet{jin2024analyzing}. And the FOL and PST representations were generated using the same method as described above.

\paragraph{WMT16, LOGIC, Pubmed45, SPIDER}
The data collection for these tasks followed the same procedure as PAWS.

\subsection{The Process of Constructing Datasets for All Tasks of SR-LLM~(training-dependent)}
\label{app:dataset_td}
In this section, I will explain the process of collecting both training and test data for the 10 tasks used in SR-LLM (training-dependent), including the original text and three types of structured representations. Data statistics are summarized in the Table~\ref{tab:dataset_of_td}.

\setlength{\tabcolsep}{4.5pt}
\renewcommand{\arraystretch}{1.2}
\begin{table}[!ht]
\centering
\caption{Tasks and datasets used in SR-LLM (training-dependent)}
\label{tab:dataset_of_td}
\small
\resizebox{\linewidth}{!}
{ 
\begin{tabular}{lccc}
\toprule
\textbf{Dataset} & \textbf{Task} & \textbf{Train Size} & \textbf{Test Size} \\
\midrule
PAWS      & Paraphrase Detection           & 10000  & 8000  \\
SNLI      & Recognizing Textual Entailment & 10000  & 10000 \\
WMT16     & Translation                    & 10000  & 5999  \\
CoNLL2003 & Named Entity Recognition       & 10000  & 3453  \\
LOGIC     & Logical Fallacy Detection      & 10000  & 2449  \\
SST-2     & Sentiment Analysis             & 10000  & 872   \\
Pubmed45  & Event Extraction               & 10000  & 5000  \\
WiC       & Lexical Disambiguation         & 5066   & 1048  \\
SPIDER    & Text2SQL Code Generation       & 7000   & 1034  \\
AGNEWS    & Text Classification            & 10000  & 7600  \\
\bottomrule
\end{tabular}
} 
\end{table}

\paragraph{PAWS, WMT16, Pubmed45, SNLI, CoNLL2003, SST-2, AGNEWS}
These datasets contain relatively large training sets. Therefore, we randomly selected 10,000 examples from each as the training set. The structured representations were generated using GPT-4o-turbo in a few-shot setting, with sample prompts provided in the figure. The test sets are the same as those used in the SR-LLM (training-free) experiments.

\begin{figure}[ht]
\centering
\vspace{0in}
\includegraphics[width=1\linewidth]{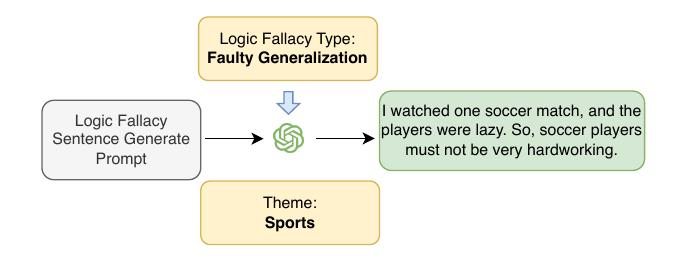}
\caption{\textbf{The synthetic process for LOGIC data.} Taking the ``Faulty Generalization'' type as an example, we employed a few-shot strategy to guide the model in generating sentences containing the logical fallacy of ``Faulty Generalization'' To ensure greater sentence diversity, we incorporated a thematic element during generation, such as ``Sports'' as shown in the figure. This thematic addition helps produce a broader variety of sentence while maintaining the specific logical error, leading to a richer and more varied dataset.}
\label{fig:process_of_making_logic}
\vspace{-0.1in}
\end{figure}

\begin{figure}[ht]
\centering
\vspace{0in}
\includegraphics[width=0.8\linewidth]{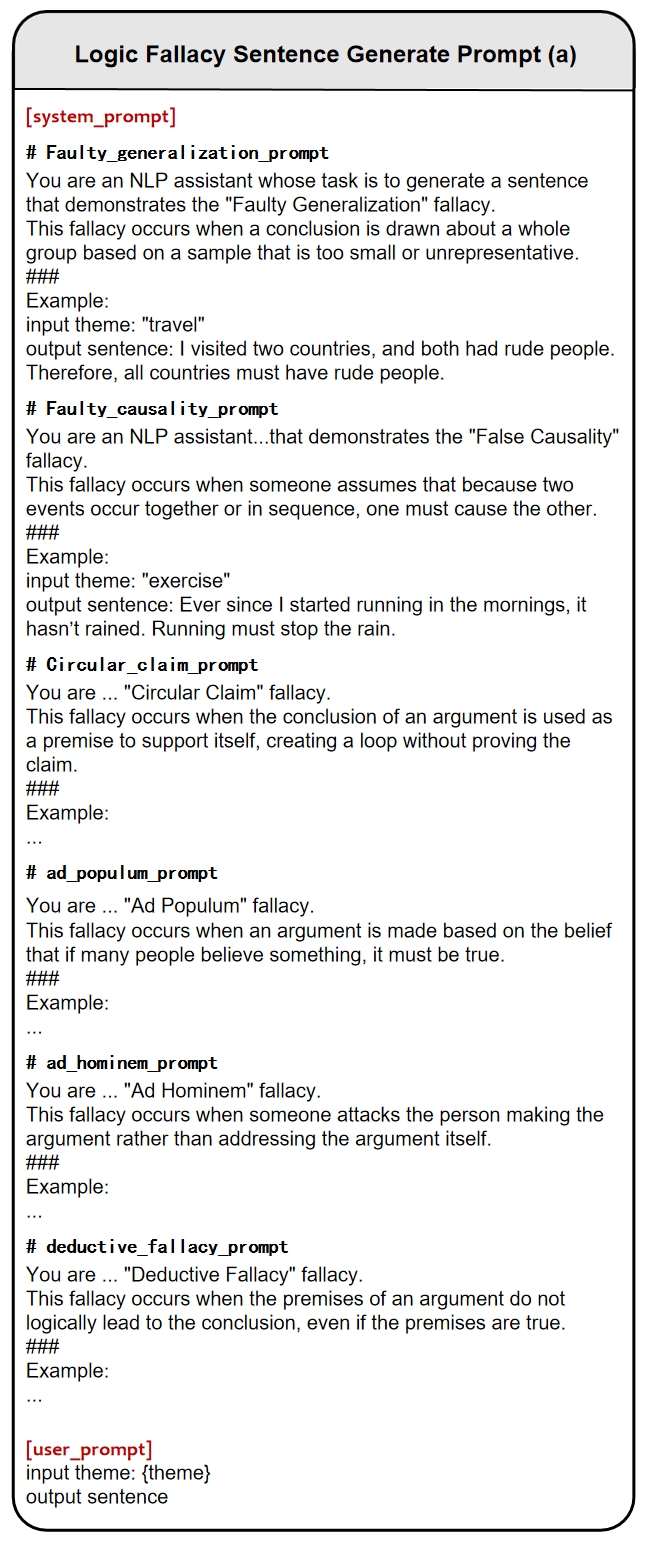}
\caption{Logic Fallacy Generate Prompt (a)}
\label{fig:makr_lfs_a}
\vspace{-0.1in}
\end{figure}

\begin{figure}[ht]
\centering
\vspace{0in}
\includegraphics[width=0.8\linewidth]{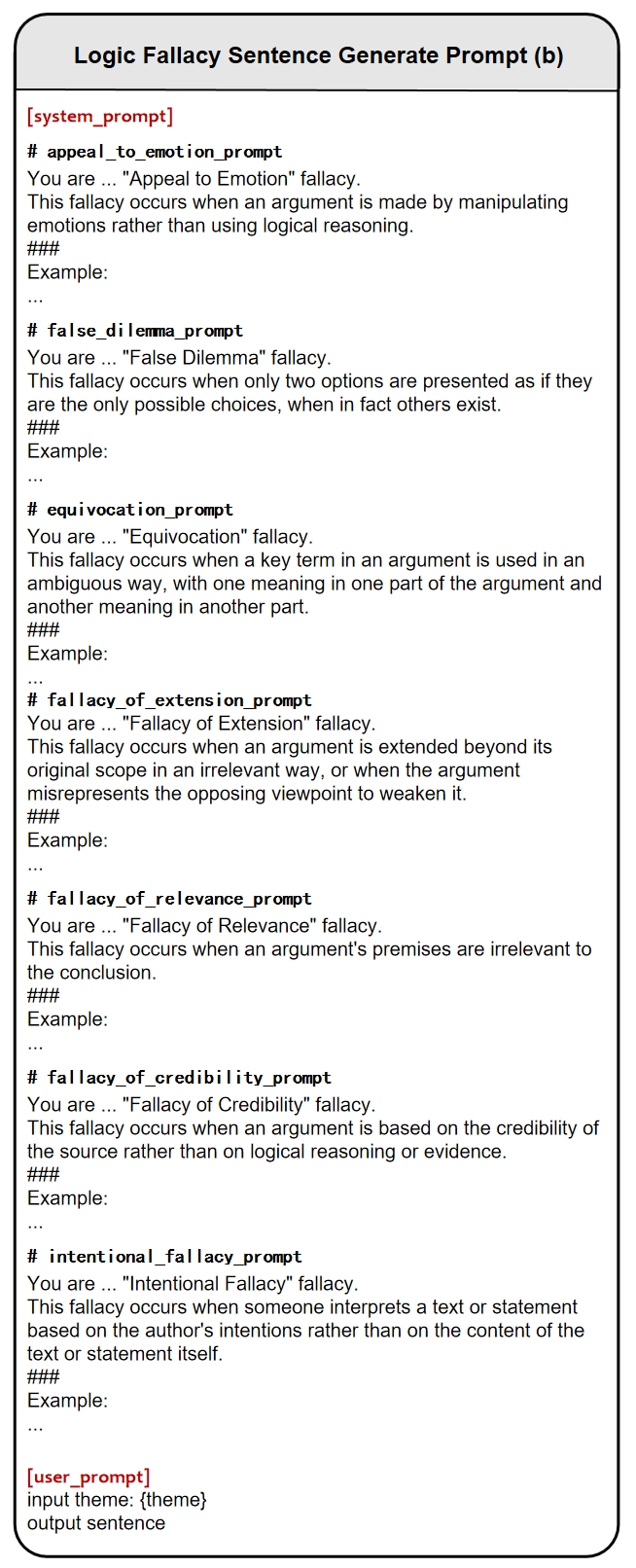}
\caption{Logic Fallacy Generate Prompt (b)}
\label{fig:makr_lfs_b}
\vspace{-0.1in}
\end{figure}

\paragraph{LOGIC}
Since the LOGIC dataset is relatively small, the training-free setup used all the available samples from the test, validation, and training sets combined, yielding a total of 2,449 samples as the test set. We retained these 2,449 samples for the test set in the training-dependent setting as well. To create the training set, we synthetically generated 10,000 logic examples using GPT-4o-turbo. The generation process is illustrated in the Figure~\ref{fig:process_of_making_logic}, where a few-shot strategy was employed to guide the model to generate sentences containing different logical fallacies. The generated prompt is shown in Figures~\ref{fig:makr_lfs_a} and Figures~\ref{fig:makr_lfs_b}. The type of logical error serves as the label, producing complete data points. Structured representations were generated in the same manner as described above.

\paragraph{SPIDER}
Since the official SPIDER test set is not publicly available, the training-free setup used a combination of training and validation sets as the test set. However, due to the complexity of generating SPIDER-like data, we used the original 7,000 training examples for the training set in the training-dependent setting and the 1,034 validation examples as the test set. Structured representations were generated as described above.

\paragraph{WiC}
As the WiC training set is relatively small, we combined the 648 validation examples with the original training set to create a total of 5,066 training samples. Structured representations were generated using the same method as described above.

\section{Additional Experiments}
\subsection{Comparative Analysis of Different SR Combinations and Their Impact on LLM Reasoning}
\begin{figure}[!ht]
\centering
\vspace{0in}
\includegraphics[width=1\linewidth]{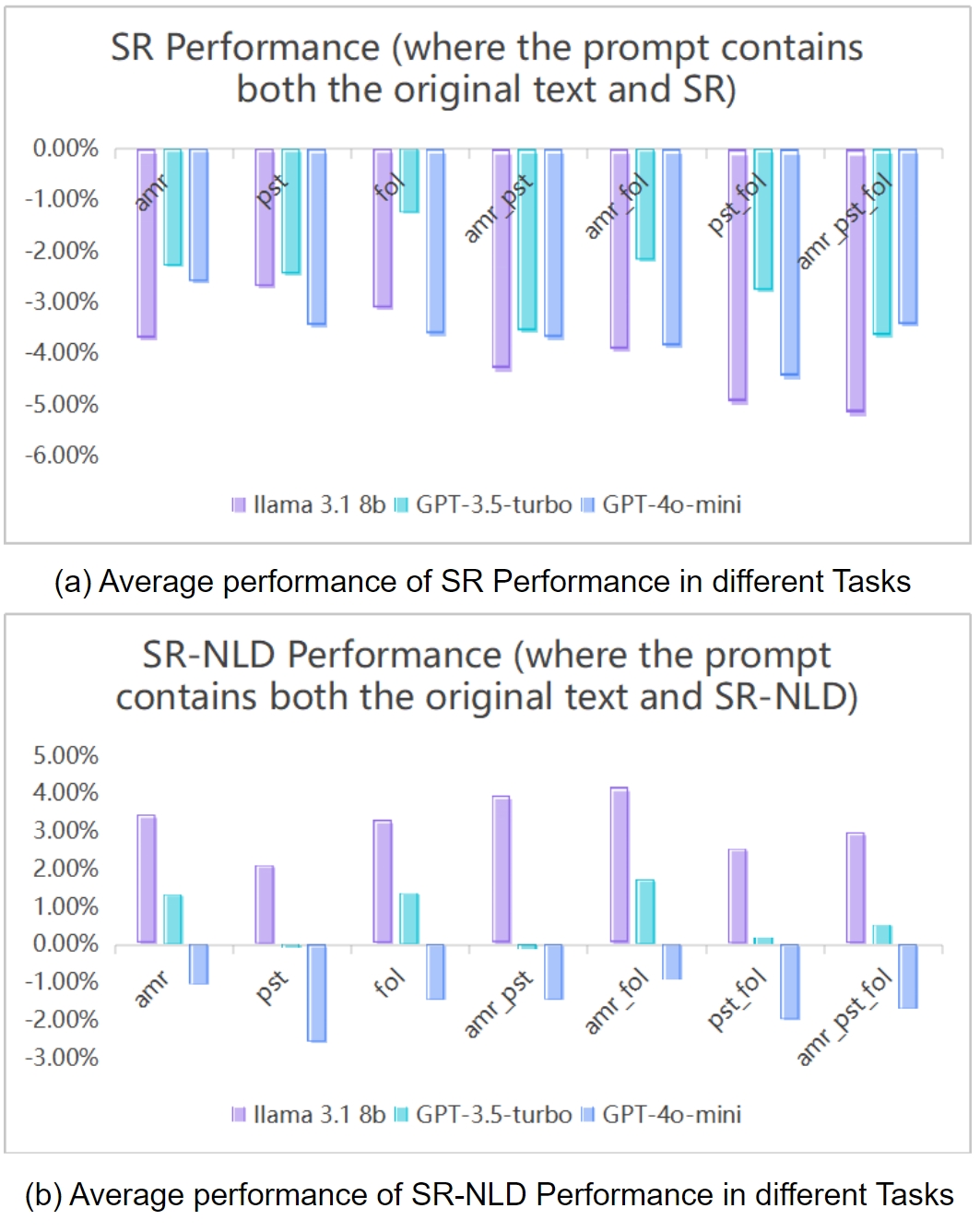}
\caption{\textbf{Performance comparison of different SR combinations.} (a) The average performance enhancement $(\Delta)$, for various SR combinations across different tasks. (b) The average performance enhancement $(\Delta)$, for different SR-NLD combinations across various tasks.}
\label{fig:performance_of_different_sr}
\end{figure}

We conducted an in-depth comparison of the performance of different structured representations (SR) and explored their combinations to assess whether joint usage could further enhance LLM reasoning capabilities. Figure \ref{fig:performance_of_different_sr} summarizes the average performance improvements across all tasks. The results indicate that the use of individual SRs such as AMR, PST, and FOL did not lead to significant performance enhancements, which is consistent with the findings of \cite{jin2024analyzing}. Moreover, when multiple SRs were introduced simultaneously, their combined complexity posed additional challenges for the LLMs, further dispersing the model’s attention and resulting in poorer performance compared to using a single SR. In contrast, when relatively weaker LLMs were provided with more comprehensible semantic features (AMR) and logical features (FOL), their average performance improved. The integration of these two types of features complemented each other, leading to better overall results. However, the contribution of syntactic features (PST) was relatively less effective and, in some cases, even negated the positive effects of semantic and logical features.

\subsection{Optimal Text-to-SR Ratio Analysis}
\label{app:ratio_select}
\begin{figure}[!ht]
\centering
\includegraphics[width=\linewidth]{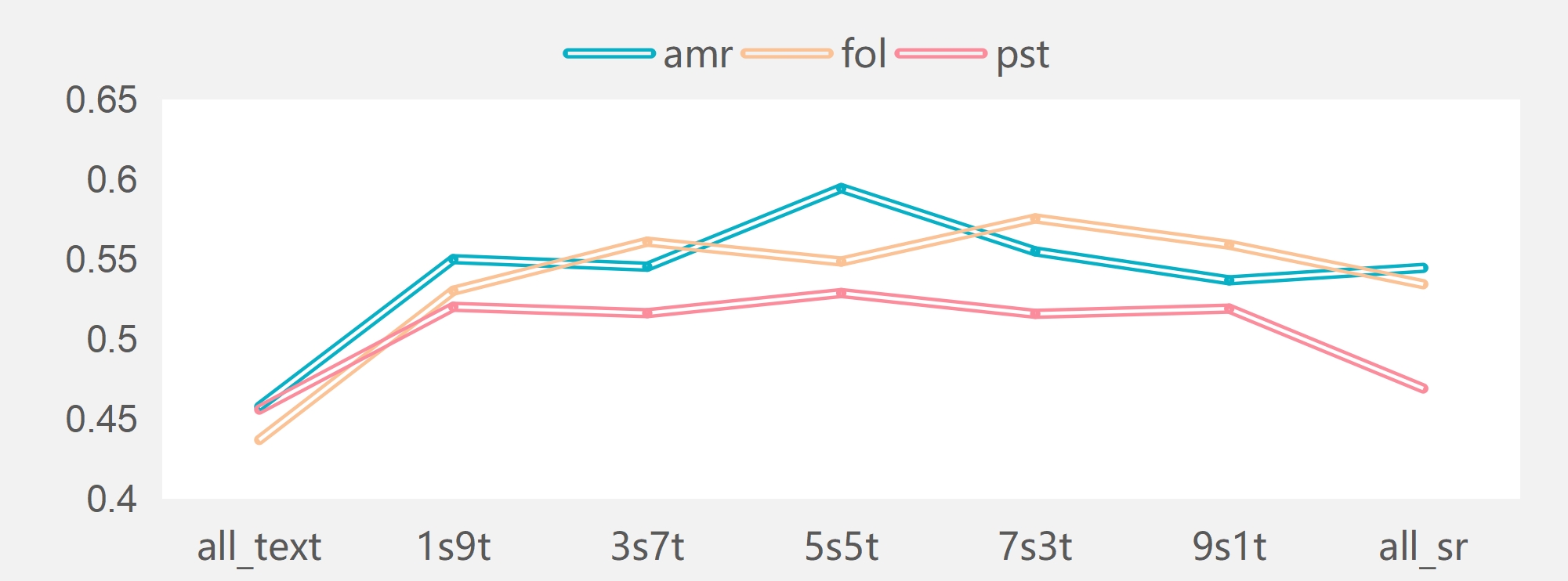}
\caption{Comparison of average performance of models at different scales in all tasks.}
\label{fig:dif_weight}
\vspace{-0.1in}
\end{figure}

To further investigate the most optimal ratio of between G(text) and G(SR), I selected five tasks, which includes PAWS, LOGIC, Pubmed45, SPIDER, WMT16 for additional experiments, adjusting the ratio of text to structured representations in the Gen-SR dataset to identify the optimal balance. The experimental results are shown in the Figure~\ref{fig:dif_weight}. As can be observed, the fluctuations in performance with different ratios are relatively small. For both AMR and PST, a 50-50 ratio between text and structured representations appears to be the most effective. However, for FOL, a 30-70 ratio (whether favoring structured representations) yields better results. This is a preliminary exploration, and I believe it represents a promising direction for further research.

\subsection{Enhancing LLM's Understanding of SR during Pretraining.}
We further conducted experiments during the pretraining phase with the goal of enhancing LLM's ability to comprehend structured representations, aiming for performance improvements in downstream tasks. Specifically, we collected 1GB of task-agnostic SR data, including AMR, PST, and FOL, following a similar procedure as in previous data collection efforts, and applied this data to the pre-training of the Llama3.1-8B-Instruct model. Building on this, we further conducted SFT, the same as SR-LLM (training-dependent), on five datasets. The final average performance results are shown in the Table~\ref{tab:model_results}.

The experimental results show that, compared to the vanilla model without pre-training, the pre-trained model indeed exhibited performance improvements in downstream tasks, though the improvements were relatively modest, with an average increase of less than 1\%. However, after applying SFT on the pre-trained model, its performance was actually inferior to that of the vanilla model trained directly with SFT. We hypothesize that this phenomenon may be due to the model forming certain inherent understandings of structured representations during the pre-training phase, which hindered its ability to establish effective connections between structure and tasks during SFT, leading to worse performance compared to the vanilla model. This phenomenon highlights a potential conflict in how the model processes structured information during the pre-training and fine-tuning phases, which warrants further exploration and resolution in future research.

\setlength{\tabcolsep}{4.5pt}
\renewcommand{\arraystretch}{1.2}
\begin{table}[!ht]
\centering
\vspace{0.1in}
\caption{\textbf{The SR enhancement of models with different training strategies.} These are the average SR Enhancement results across all tasks under different training strategies. Green indicates the best performance within the same SR, while red represents the worst performance.}
\label{tab:model_results}
\vspace{-0.1in}
\footnotesize
\resizebox{0.5\textwidth}{!}
{ 
\begin{tabular}{lcccc|c}
\toprule
AMR & FOL & PST & Pretrain & SFT & $\Delta$ \\
\midrule
\checkmark   &     &     &          &     & {\color[HTML]{FE0000} \textbf{-3.51\%}} \\
\checkmark   &     &     & \checkmark        &     & 0.56\%                                  \\
\checkmark   &     &     & \checkmark        & \checkmark   & -1.16\%                                 \\
\checkmark   &     &     &          & \checkmark   & {\color[HTML]{34FF34} \textbf{11.59\%}} \\
\midrule
    & \checkmark   &     &          &     & {\color[HTML]{FE0000} \textbf{-2.83\%}} \\
    & \checkmark   &     & \checkmark        &     & 1.30\%                                  \\
    & \checkmark   &     & \checkmark        & \checkmark   & 3.10\%                                  \\
    & \checkmark   &     &          & \checkmark   & {\color[HTML]{34FF34} \textbf{6.45\%}}  \\
\midrule
    &     & \checkmark   &          &     & {\color[HTML]{FE0000} \textbf{-3.61\%}} \\
    &     & \checkmark   & \checkmark        &     & -0.18\%                                 \\
    &     & \checkmark   & \checkmark        & \checkmark   & 1.58\%                                  \\
    &     & \checkmark   &          & \checkmark   & {\color[HTML]{34FF34} \textbf{2.91\%}} \\
\bottomrule
\end{tabular}
} 
\vspace{-0.1in}
\end{table}

\section{Examples of Gen-SR}
\label{app:prompt_gensr}
\begin{figure}[!ht]
\centering
\vspace{0.1in}
\includegraphics[width=0.8
\linewidth]{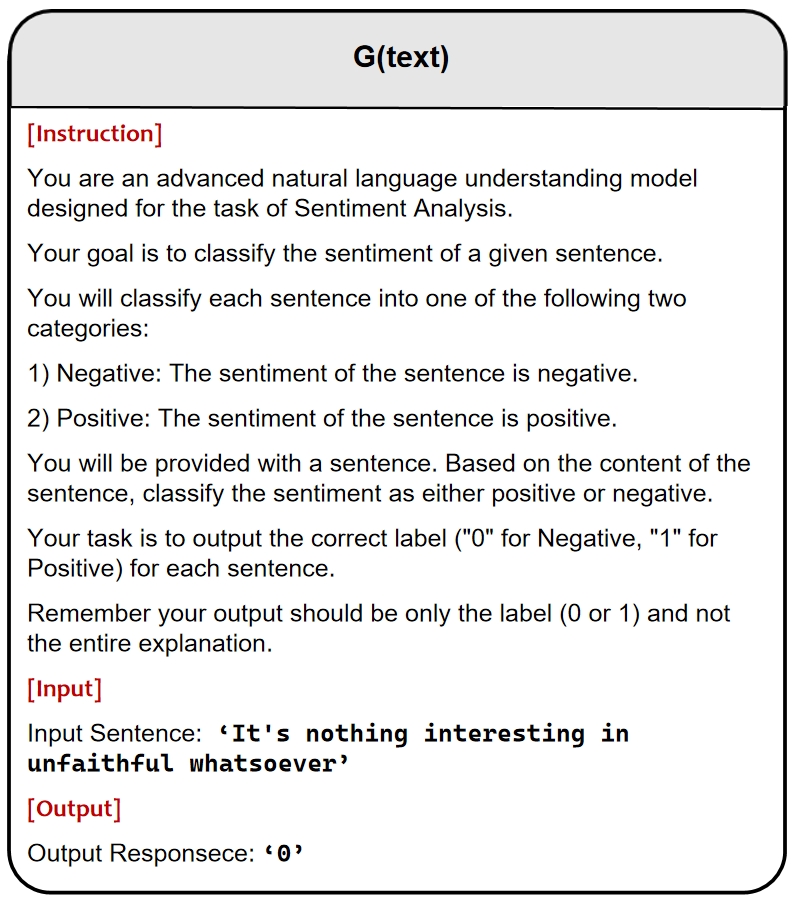}
\vspace{0.1in}
\caption{The Example of G(text)}
\label{fig:g_text}
\vspace{0.1in}
\end{figure}

\begin{figure}[!ht]
\centering
\vspace{0.1in}
\includegraphics[width=0.8
\linewidth]{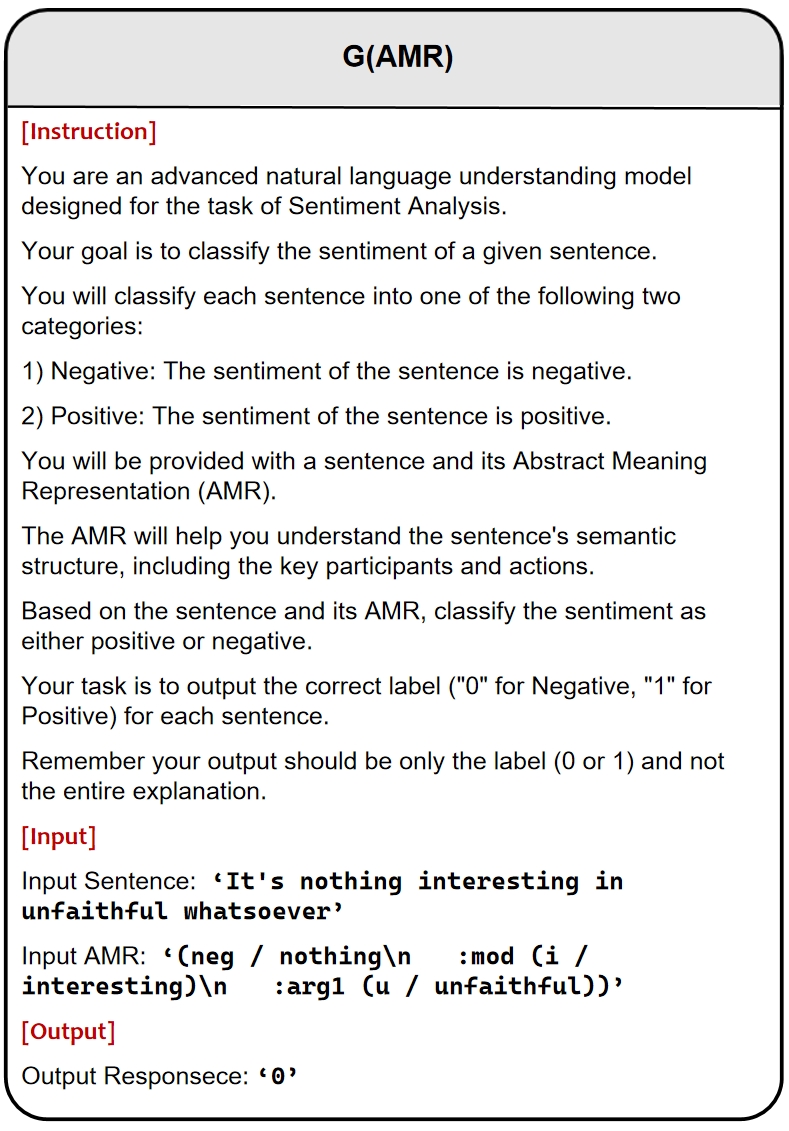}
\vspace{0.1in}
\caption{The Example of G(AMR)}
\label{fig:g_amr}
\vspace{0.1in}
\end{figure}

\begin{figure}[!ht]
\centering
\vspace{0.1in}
\includegraphics[width=0.8
\linewidth]{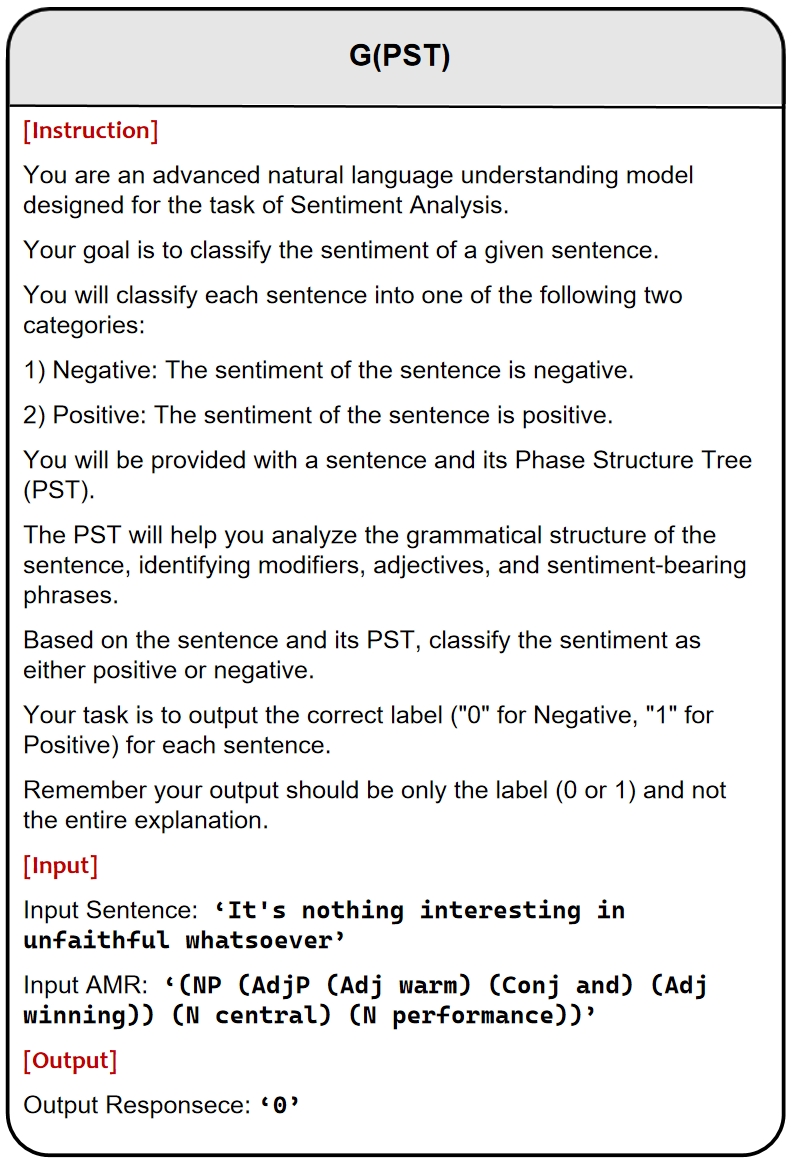}
\vspace{0.1in}
\caption{The Example of G(PST)}
\label{fig:g_pst}
\vspace{0.1in}
\end{figure}

\begin{figure}[!ht]
\centering
\vspace{0.1in}
\includegraphics[width=0.8
\linewidth]{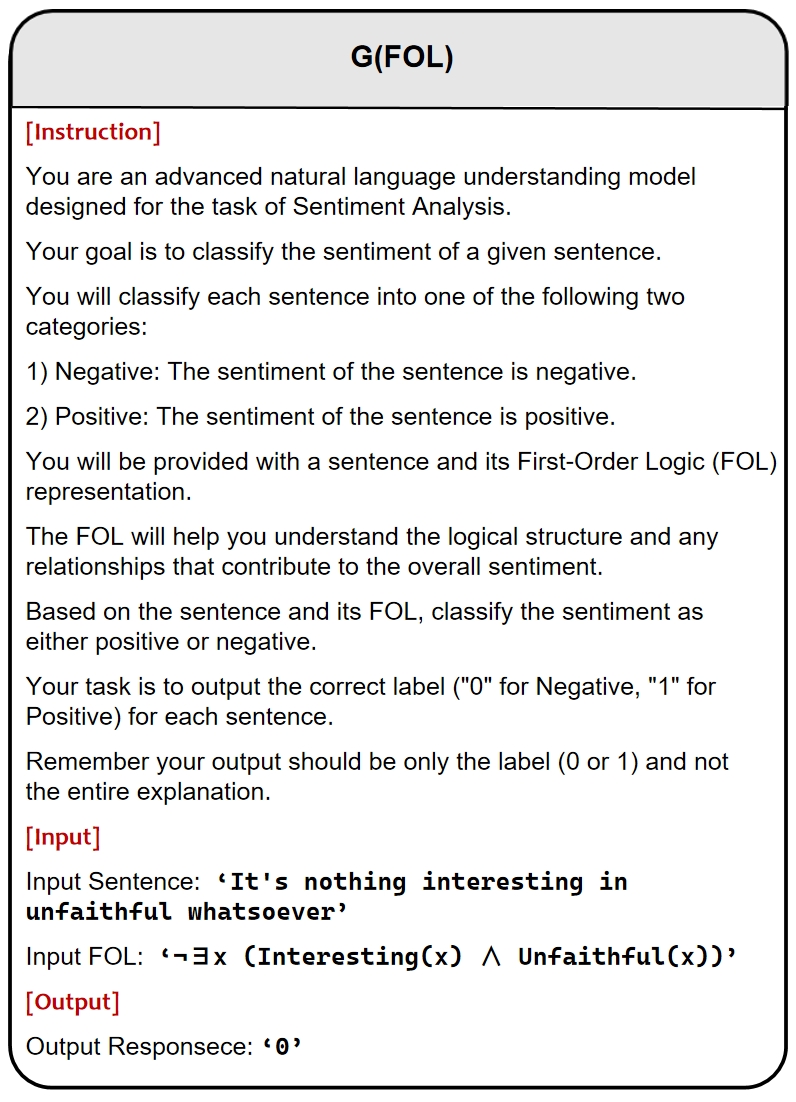}
\vspace{0.1in}
\caption{The Example of G(FOL)}
\label{fig:g_fol}
\vspace{0.1in}
\end{figure}

We present specific examples of Gen-SR in this section. Figure~\ref{fig:g_text} shows an example of G(text), Figure~\ref{fig:g_amr} shows an example of G(AMR), Figure~\ref{fig:g_pst} shows an example of G(PST), and Figure~\ref{fig:g_fol} shows an example of G(FOL).

\section{Prompt of Testing the SR-LLM}
\label{app:prompt}
\begin{figure}[!ht]
\centering
\vspace{0.1in}
\includegraphics[width=0.8
\linewidth]{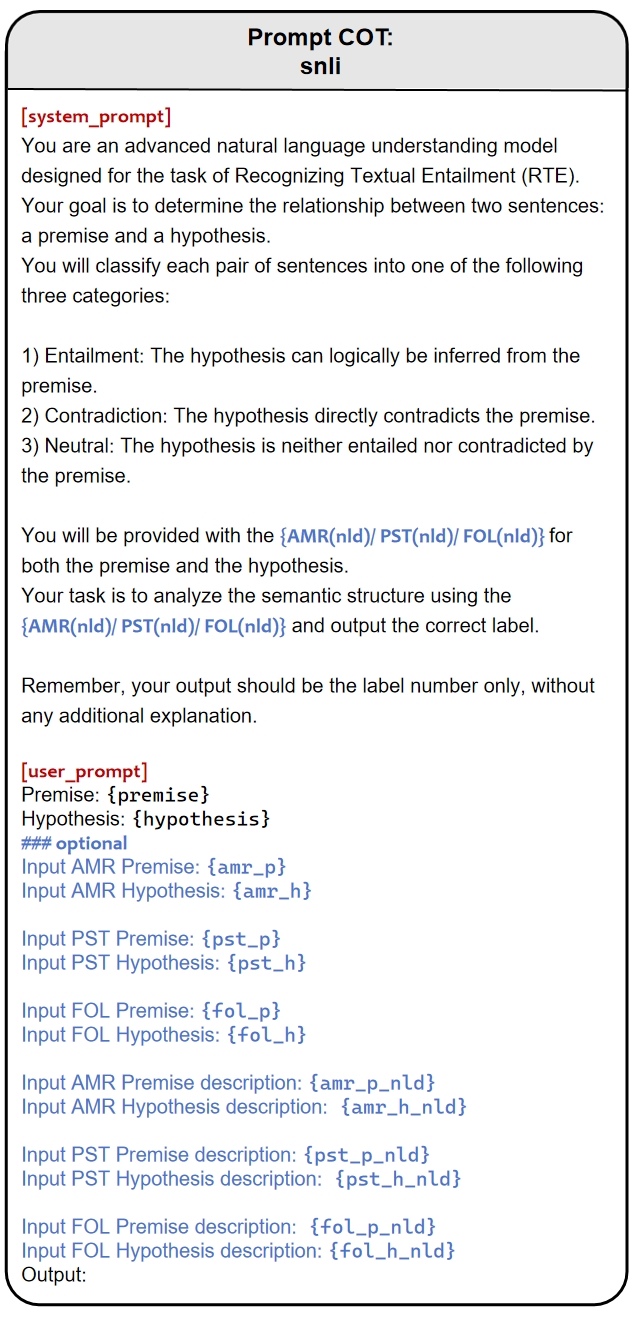}
\vspace{0.1in}
\caption{The COT prompt of SNLI}
\label{fig:cot_snli}
\vspace{0.1in}
\end{figure}

\begin{figure}[!ht]
\centering
\vspace{0.1in}
\includegraphics[width=0.8
\linewidth]{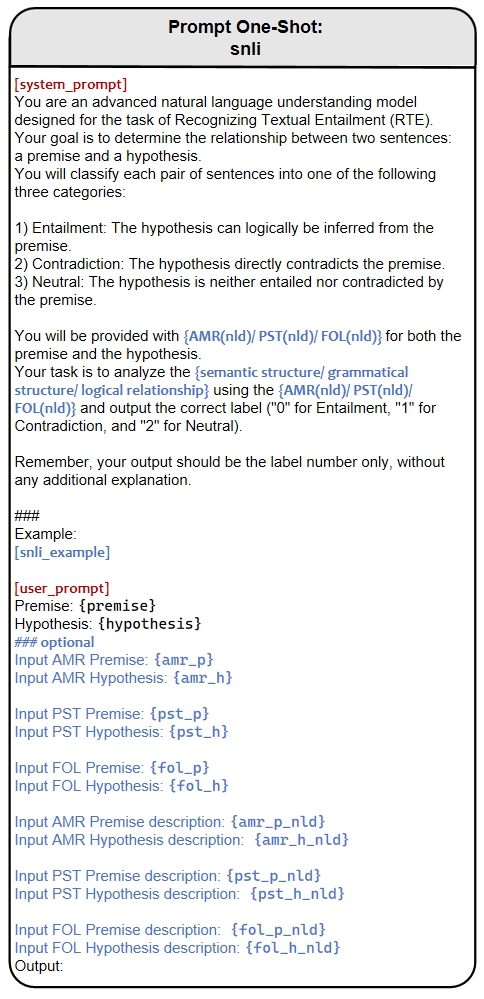}
\vspace{0.1in}
\caption{The One-Shot prompt of SNLI}
\label{fig:oneshot_snli}
\vspace{0.1in}
\end{figure}

\begin{figure}[!ht]
\centering
\vspace{0.1in}
\includegraphics[width=0.8
\linewidth]{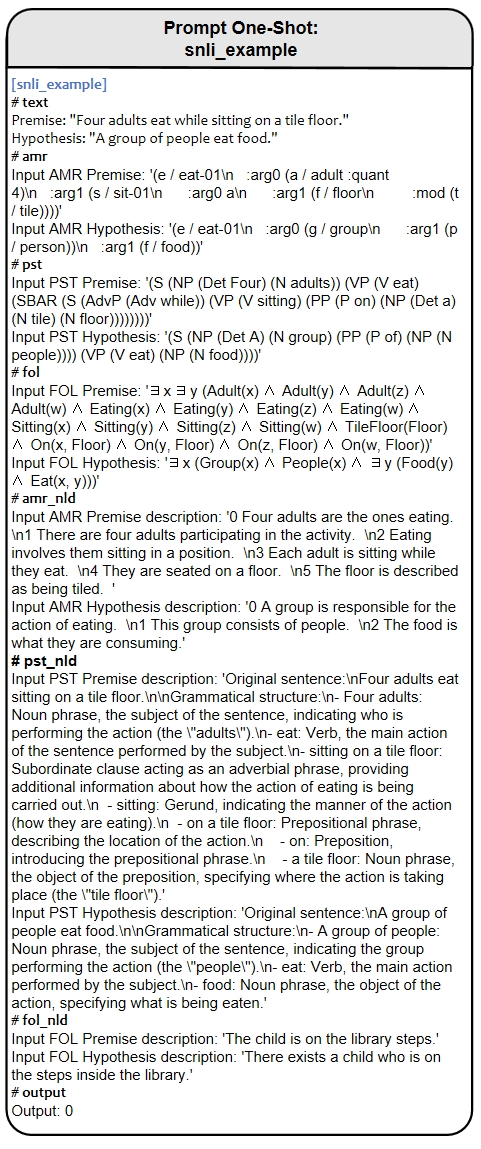}
\vspace{0.1in}
\caption{The One-Shot prompt of SNLI's example}
\label{fig:oneshot_snli_e}
\vspace{0.1in}
\end{figure}

We present the complete prompts for our experiments, including both CoT and One-shot examples, using the SNLI dataset as an illustration in Figures~\ref{fig:cot_snli}, Figures~\ref{fig:oneshot_snli} and Figures~\ref{fig:oneshot_snli_e}.






\end{document}